\documentclass[runningheads]{llncs}

\PassOptionsToPackage{numbers, compress}{natbib}
\usepackage[numbers,square,sort]{natbib}

\setcitestyle{}
\usepackage[utf8]{inputenc}
\usepackage{amsmath}
\usepackage{amssymb}
\usepackage{color}
\usepackage{xifthen}
\usepackage{amsxtra}
\usepackage{xspace}
\usepackage[ruled, vlined, linesnumbered]{algorithm2e}
\usepackage{etex}
\usepackage{cvpr-abbriv}
\usepackage{array}
\usepackage{url}
\usepackage{graphicx}
\usepackage{xifthen}
\usepackage[labelfont=bf]{caption}
\makeatletter

\let\proof\@undefined                        % undefine \proof
\let\endproof\@undefined                  % undefine \endproof
\makeatother
\usepackage{amsmath,amssymb,bm,amsthm,amsxtra,stmaryrd,thmtools, thm-restate}
\usepackage{chngcntr}
\usepackage{tocloft}
\usepackage{booktabs}
\usepackage{xcolor}
\usepackage{listings}
\definecolor{dkgreen}{rgb}{0,0.6,0}
\definecolor{gray}{rgb}{0.5,0.5,0.5}
\definecolor{mauve}{rgb}{0.58,0,0.82}
\usepackage[most]{tcolorbox}
\usepackage{pbox}
\newlength{\luw}
\newlength{\luh}

\usepackage{tikz}% create images using pgf/tikz

\usepackage{atbeginend}
\usepackage{hyperref}
\hypersetup{draft=false, pagebackref=false,breaklinks=true,colorlinks,bookmarks=false,pdftex,hidelinks,colorlinks=false,linkbordercolor={1 1 1}}
\hypersetup{bookmarksopen=true} % expand table of contents in the viewer by default

\def\E{{\mathbb{E}}}

\def\d{\mathrm{d}}
\def\eps{\varepsilon}
\renewcommand{\Pr}{\mathbb{P}}
\newcommand{\Real}{\mathbb{R}}

\newcommand{\V}{\mathbb{V}}
\newcommand{\U}{\mathcal{U}}
\newcommand{\sigmoid}{\sigma}

\newcommand{\gray}{\color[rgb]{0.5,0.5,0.5}}
\newcommand{\red}{\color[rgb]{1,0,0}}

\DeclareMathOperator*{\argmax}{arg\,max}

\def\logit{{\rm logit}}
\DeclareMathOperator*{\sign}{sign}

\newcommand{\revisit}[1][]{%
\ifthenelse{\equal{#1}{}}{% no argument
\ensuremath{\red \triangle}\xspace}{%
{\ensuremath{\red \rhd}\xspace}%
{\gray #1}%
{\ensuremath{\red \lhd}\xspace}%
}%
}

\newcolumntype{L}[1]{>{\raggedright\let\newline\\\arraybackslash\hspace{0pt}}m{#1}}
\newcolumntype{C}[1]{>{\centering\let\newline\\\arraybackslash\hspace{0pt}}m{#1}}
\newcolumntype{R}[1]{>{\raggedleft\let\newline\\\arraybackslash\hspace{0pt}}m{#1}}
\newcolumntype{P}[1]{>{\centering}p{#1}}



\def\mathrlap{\mathpalette\mathrlapinternal} 
\def\mathllap{\mathpalette\mathllapinternal}
\def\mathllapinternal#1#2{\llap{$\mathsurround=0pt#1{#2}$}}
\def\mathrlapinternal#1#2{\rlap{$\mathsurround=0pt#1{#2}$}}

\def\leftbb{\mathrlap{[}\hskip1.3pt[}
\def\rightbb{]\hskip1.36pt\mathllap{]}}

\makeatletter
\newcommand*{\rom}[1]{{\expandafter\@slowromancap\romannumeral #1@}}
\newcommand*{\Rom}[1]{{\bf \expandafter\@slowromancap\romannumeral #1@)}}
\makeatother

\newlength{\myskip}
\SetAlgoSkip{skipalgomyskip}
\SetAlgoInsideSkip{}
\makeatletter
\let\corollary\@undefined
\let\c@corollary\@undefined
\let\endcorollary\@undefined
\let\definition\@undefined
\let\c@definition\@undefined
\let\enddefinition\@undefined
\let\proof\@undefined
\let\endproof\@undefined
\let\theorem\@undefined
\let\c@theorem\@undefined
\let\endtheorem\@undefined
\let\lemma\@undefined
\let\c@lemma\@undefined
\let\endlemma\@undefined
\let\example\@undefined
\let\c@example\@undefined
\let\endexample\@undefined
\let\remark\@undefined
\let\c@remark\@undefined
\let\endremark\@undefined
\let\proposition\@undefined
\let\c@proposition\@undefined
\let\endproposition\@undefined
\let\property\@undefined
\let\endproperty\@undefined
\makeatother

\usepackage{thmtools}

\newtheoremstyle{tightItalic}% name
  {0.5\myskip}%      Space above
  {0\myskip}%      Space below
  {}%         Body font
  {}%         Indent amount (empty = no indent, \parindent = para indent)
  {\itshape}% Thm head font
  {.}%        Punctuation after thm head
  { }%     Space after thm head: " " = normal interword space;
  {}%         Thm head spec (can be left empty, meaning `normal')

\newtheoremstyle{TheoremStyle}% name
  {0.5\myskip}%      Space above
  {0.5\myskip}%      Space below
  {\itshape}%         Body font
  {}%         Indent amount (empty = no indent, \parindent = para indent)
  {\bf}% Thm head font
  {.}%        Punctuation after thm head
  { }%     Space after thm head: " " = normal interword space;
  {}%         Thm head spec (can be left empty, meaning `normal')

\newtheoremstyle{tightBf}% name
  {0.5\myskip}%      Space above
  {0\myskip}%      Space below, 0 since amsart starts a paragraph
  {}%         Body font
  {}%         Indent amount (empty = no indent, \parindent = para indent)
  {\bf}% Thm head font
  {.}%        Punctuation after thm head
  {.5em}%     Space after thm head: " " = normal interword space;
  {}%         Thm head spec (can be left empty, meaning `normal')

\theoremstyle{definition}
\theoremstyle{tightBf}

\declaretheorem[style=TheoremStyle,name=Theorem]{theorem}
\declaretheorem[style=TheoremStyle,name=Lemma]{lemma}
\declaretheorem[style=tightBf,name=Definition]{definition}
\declaretheorem[style=TheoremStyle,name=Corollary]{corollary}
\declaretheorem[style=TheoremStyle,name=Proposition]{proposition}

\declaretheorem[style=tightBf,name=Example]{example}
\declaretheorem[style=tightItalic,name=Remark]{remark}
\theoremstyle{tightItalic}
\newtheorem*{proof}{Proof}
\BeforeEnd{proof}{\qed}


\usepackage[capitalise,nameinlink]{cleveref}
\crefformat{equation}{(#2#1#3)}
\crefname{observation}{Observation}{Observations}
\crefname{algorithm}{Alg.}{Algs.}
\let\displaystyle\textstyle
\renewcommand{\paragraph}[1]{\subsubsection{#1}}

\def\mytitle{Bias-Variance Tradeoffs in Single-Sample Binary Gradient Estimators}
\def\AS{Alexander Shekhovtsov}
\def\CTU{Czech Technical University in Prague}
\def\ASemail{shekhole.fel.cvut.cz}
\def\gst{\hat g_{\text{\sc st}}}
\def\gdarn{\hat g_{\text{\sc darn}}}

\def\ggst{\hat g_{\text{\sc gs}(\tau)}}
\DeclareMathOperator*{\Bernoulli}{Bernoulli}
\DeclareMathOperator*{\Bin}{Bin}
\def\f{f}
\def\SC#1{\text{\uppercase{#1}}\xspace}
\def\GS{\SC{gs}}

\def\ST{\SC{st}}
\def\DARN{\SC{darn}}
\def\REINFORCE{{\sc reinforce}\xspace}
\newif\ifreview
\reviewfalse

\ifreview
	\usepackage{lineno}

	\linenumbers
\fi

\begin{document}

%%%%%%%%%%%%%%%%%%%%% Add submission id, track, and title. %%%%%%%%%%%%%%%%%%%%%

% Insert your submission number here
\def\SubNumber{036}

% Choose one track by uncommenting one of the following lines  
\def\GCPRTrack{Regular Track}
% \def\GCPRTrack{Track: Computer vision systems and applications}
% \def\GCPRTrack{Track: Pattern recognition in the life and natural sciences}
% \def\GCPRTrack{Track: Photogrammetry and remote sensing}
% \def\GCPRTrack{Track: Robot vision}
% \def\GCPRTrack{Track: DAGM Young Researcher Forum}

% Replace with your title
\title{\mytitle\thanks{The author gratefully acknowledges support by Czech OP VVV project ``Research Center for Informatics (CZ.02.1.01/0.0/0.0/16019/0000765)''}}
% You can use \thanks for acknowledgment. Do not add any acknowledgment to the draft 
% version that is used for the review process.  
%\title{Title\thanks{XXX}}

\ifreview
	% ANONYMOUS SUBMISSION FOR REVIEW
	% DO NOT MODIFY these for the draft version that is used for the review process.
	\titlerunning{DAGM GCPR 2021 Submission \SubNumber{}. CONFIDENTIAL REVIEW COPY.}
	\authorrunning{DAGM GCPR 2021 Submission \SubNumber{}. CONFIDENTIAL REVIEW COPY.}
	\author{DAGM GCPR 2021 - \GCPRTrack{}}
	\institute{Paper ID \SubNumber}
\else
	\titlerunning{}
	%\titlerunning{Bias-Variance Tradeoffs in Recent Single-Sample Binary Gradient Estimators}

	\authorrunning{A. Shekhovtsov}
	%\authorrunning{~}
	\author{\AS\\ %\orcidID{0000-1111-2222-3333}\\
	\email{\ASemail}
	\institute{\CTU}}

\fi

\maketitle              % typeset the header of the contribution
%%%%%%%%%%%%%%%%%%%%%%%%%%%%%%END TEMPLATE %%%%%%%%%%%%%%%%%%%%%%%%%%%%%%%%%%

\begin{abstract}
Discrete and especially binary random variables occur in many machine learning models, notably in variational autoencoders with binary latent states and in stochastic binary networks. When learning such models, a key tool is an estimator of the gradient of the expected loss with respect to the probabilities of binary variables. The straight-through (ST) estimator gained popularity due to its simplicity and efficiency, in particular in deep networks where unbiased estimators are impractical. Several techniques were proposed to improve over ST while keeping the same low computational complexity: Gumbel-Softmax, ST-Gumbel-Softmax, BayesBiNN, FouST. 
We conduct a theoretical analysis of bias and variance of these methods in order to understand tradeoffs and verify the originally claimed properties. The presented theoretical results allow for better understanding of these methods and in some cases reveal serious issues.
\end{abstract}

\section{Introduction}
Binary variables occur in many models of interest. Variational autoencoders (VAE) with binary latent states are used to learn generative models with compressed representations~\cite{Gregor-14,MuProp,Pervez-20,Vahdat2020PMLR} and to learn binary hash codes for text and image retrieval~\cite{Chaidaroon-17,Shen-18,Dadaneh2020PairwiseSH,Nanculef-20}. Neural networks with binary activations and weights are extremely computationally efficient and attractive for embedded applications, in particular pushed forward in the vision research~\cite{Horowitz-14,Esser-16,rastegari2016xnor,Bulat_2017_ICCV,Xiang-17,zhou2016dorefa,alizadeh2018a,liu2018bi,Bethge-19,Tang2017HowTT,bulat2019improved,Martinez20,bulat2021}. Training these discrete models is possible via the stochastic relaxation, equivalent to training a Stochastic Binary Networks (SBN)~\cite{peters2018probabilistic,Raiko-14,PSA,Roth-19,shayer2018learning}. In this relaxation, each binary weight is replaced with a Bernoulli random variable and each binary activation is replaced with a conditional Bernoulli variable. The gradient of the expected loss in the weight probabilities is well defined and SGD optimization can be applied. %Further applications of binary decision variables may include conditional computation~\cite{bengio2013estimating} and structure optimization.

\begin{figure}[t]
\centering
\includegraphics[width=0.7\linewidth]{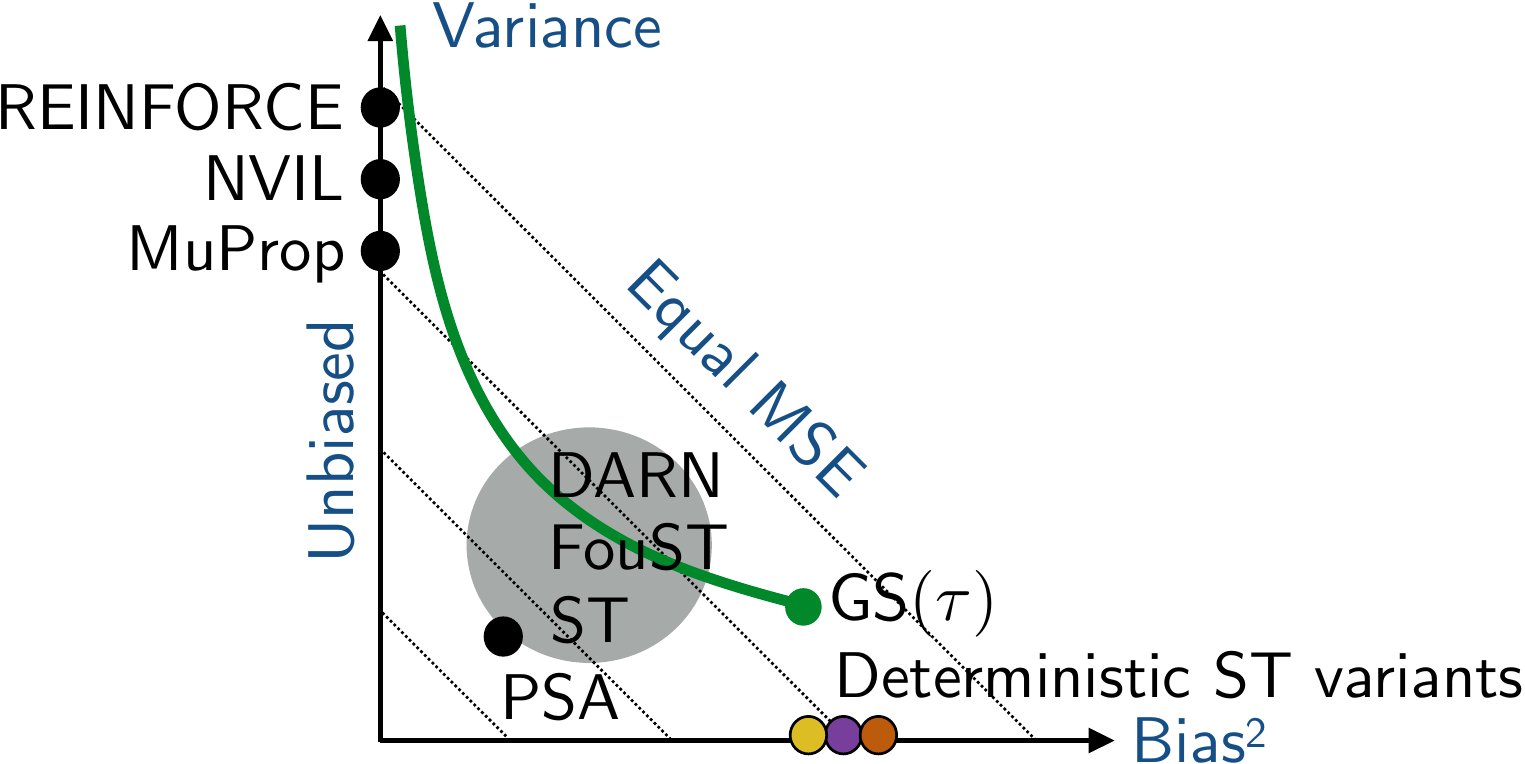}
\caption{\label{fig:tradeoffs} Schematic illustration of bias-variance tradeoffs (we do not pretend on exactness, but see experimental evaluations in~\cite{PSA,ST}; notice that the Mean Squared Error (MSE) is the sum of variance and squared bias). Unbiased methods have a prohibitively high variance for deep models. PSA achieves a significant reduction in variance at a price of a small bias, but has a limited applicability. According to~\cite{PSA}, ST estimator can be as accurate as PSA in wide deep models. We analytically study methods in the gray area: GS, DARN and FouST in order to find out whether they can offer a sound improvement over ST. In particular, for GS estimator the figure illustrates its possible tradeoffs when varying the temperature parameter according to the asymptotes we prove.
}
\end{figure}

For the problem of estimating gradient of expectation in probabilities of (conditional) Bernoulli variables, several unbiased estimators were proposed~\cite{Mnih-2014,MuProp,grathwohl18-relax,Tucker-17-REBAR,yin18-arm}. However, in the context of deep SBNs these methods become impractical: MuProp~\cite{MuProp} and \REINFORCE with baselines~\cite{Mnih-2014} have a prohibitively high variance in deep layers~\cite[Figs. C6, C7]{ST} while other methods' complexity grows quadratically with the number of Bernoulli layers. In these cases, biased estimators were more successful in practice: straight-through (\ST)~\cite{ST}, Gumbel-Softmax (\GS)~\cite{jang2016categorical,maddison2016concrete} and their variants. 
In order to approximate the gradient of the expectation these methods use a single sample of all random entities and the derivative of the objective function extended to the real-valued domain. A more accurate PSA method was presented in~\cite{PSA}, which has low computation complexity, but applies only to SBNs of classical structure\footnote{Feed-forward, with no residual connections and only linear layers between Bernoulli activations.} and requires specialized convolutions. Notably, it was experimentally reported~\cite[Fig.4]{PSA} that the baseline \ST performs nearly identically to PSA in moderate size SBNs. \cref{fig:tradeoffs} schematically illustrates the bias-variance tradeoff with different approaches. 
%Notably, \citet{PSA} found that \ST is very accurate for SBNs in practice: their more accurate PSA method, with reduced bias, performs identically to \ST~\cite[Fig.4]{PSA}.
%

\paragraph{Contribution}
In this work we analyze theoretical properties of several recent single-sample gradient based methods: \GS, \ST-\GS~\cite{jang2016categorical}, BayesBiNN~\cite{meng-20} and FouST~\cite{Pervez-20}. We focus on clarifying these techniques, studying their limitations and identifying incorrect and over-claimed results. We give a detailed analysis of bias and variance of \GS and \ST-\GS estimators. Next we analyze the application of GS in BayesBiNN. We show that a correct implementation would result in an extremely high variance. However due to a hidden issue, the estimator in effect reduces to a deterministic straight-through (with zero variance). A long-range effect of this swap is that BayesBiNN fails to solve the variational Bayesian learning problem as claimed. FouST~\cite{Pervez-20} proposed several techniques for {\em lowering} bias and variance of the baseline \ST estimator. We show that the baseline \ST estimator was applied incorrectly and that some of the proposed improvements may increase bias and or variance. %of (simpler) baseline methods and that the .
%
%We believe this analysis is very needed in the field, as these works are misleading and blocks 

We believe these results are valuable for researchers interested in applying these methods, working on improved gradient estimators or  developing Bayesian learning methods. Incorrect results with hidden issues in the area could mislead many researchers and slow down development of new methods.

\paragraph{Outline}
The paper is organized as follows. In \cref{sec:background} we briefly review the baseline ST estimator. %In section~\cref{GS} the properties of GS estimator are analysed propose analysis of 
%
%In sections~\cref{GS,BayesBiNN,FouST} we an
In the subsequent sections we analyze Gumbel-Softmax estimator (\cref{sec:GS}), BayesBiNN (\cref{sec:BayesBiNN}) and FouST estimator (\cref{sec:FouST}). 
Proofs are provided in the respective Appendices A to C.
As most of our results are theoretical, simplifying derivation or identifying limitations and misspecifications of the preceding work, we do not propose extensive experiments. Instead, we refer to the literature for the experimental evidence that already exists and only conduct specific experimental tests as necessary. 
In \cref{sec:conclusion} we summarize our findings and discuss how they can facilitate future research.
 %However we refer to the literature for the existing experimental evidence and provide additional experiments whether the results in the literature are contradictory. 

\section{Background}\label{sec:background}
%\revisit[explain $f$]
We define a stochastic binary unit $x\sim \Bernoulli(p)$ as $x=1$ with probability $p$ and $x=0$ with probability $1-p$. Let $f(x)$ be a loss function, which in general may depend on other parameters and may be stochastic aside from the dependence on $x$. This is particularly the case when $f$ is a function of multiple binary stochastic variables and we study its dependence on one of them explicitly. The goal of binary gradient estimators is to estimate
\begin{align}\label{g-def}
g = \frac{\d}{\d p}\E[f(x)],
\end{align}
where $\E$ is the total expectation. 
Gradient estimators which we consider make a stochastic estimate of the total expectation by taking a single joint sample. We will study their properties with respect to $x$ only given the rest of the sample fixed. In particular, we will confine the notion of bias and variance to the conditional expectation $\E_x$ and the conditional variance $\V_x$.
%In our analysis, we will study the properties o
%In our analysis we will %restrict ourselves to the case when these sources of randomness are independent of $x$, \eg, when $f $
%
%Let $y$ denote (a vector of) other (stochastic) units in the network, independent of $x$.
%We consider a function $f(x,y)$ defined analytically on $\Real\times \Y$. %, of which only the values on $\{-1,1\}\times \Y$ are relevant for a forward pass, but the 
We will assume that the function $f(x)$ is defined on the interval $[0,1]$ and is differentiable on this interval. This is typically the case when $f$ is defined as a composition of simple functions, such as in neural networks.
While for discrete inputs $x$, the continuous definition of $f$ is irrelevant, it will be utilized by approximations exploiting its derivatives. % of $f$.
%This extension is utilized to approximate derivatives and evaluate relaxations.

The expectation $\E_x [f(x)]$ can be written as
\begin{align}
(1-p) f(0) + p f(1),
\end{align}
%where $f(x) = \E_{y} f(x,y)$. 
%The true gradient of $\E [f(x,y)]$ 
Its gradient in $p$ is respectively
\begin{align}
g = \frac{\d}{\d p} \E_x [f(x)] = f(1) - f(0).
\end{align}
While this is simple for one random variable $x$, it requires evaluating $f$ at two points. With $n$ binary units in the network, in order to estimate all gradients stochastically, we would need to evaluate the loss $2n$ times, which is prohibitive. 

Of high practical interest are stochastic estimators that evaluate $f$ only at a single joint sample (perform a single forward pass).
Arguably, the most simple such estimator is the straight-through (\ST) estimator:
\begin{align}\label{ST}
\gst = f'(x).
\end{align}
For an in-depth introduction and more detained study of its properties we refer to~\cite{ST}. 
%Here, the factor $2$ occurs due to $\pm 1$ encoding used for $x$. Indeed, if we wished to use the Boolean encoding $b=(x+1)/2$ and define the equivalent loss function $f_{\rm b}(b) = f(2x-1)$, the \ST estimator would become $\gst = 2 f'(x) = f'(b)$.
%
%where $'$ shall refer to the derivative in the first argument.
The mean and variance of this \ST estimator are given by
\begin{subequations} 
\begin{align}
\E_x[\gst ] &= (1-p)f'(0) + pf'(1),\\
\V_x[\gst ] &= \E_x [\gst ^2] - (\E_x [\gst ])^2 = p (1-p)(f'(1) - f'(0))^2.
\end{align}
\end{subequations}
If $f(x)$ is linear in $x$, \ie, $f(x) = h x + c$, where $h$ and $c$ may depend on other variables, then $f'(0) = f'(1) = h$ and $f(1)-f(0) = h$. In this linear case we obtain
\begin{subequations} 
\begin{align}
\E_x[\gst ] &= h,\\
\V_x[\gst ] &= 0.
\end{align}
\end{subequations}
From the first expression we see that the estimator is unbiased and from the second one we see that its variance (due to $x$) is zero. It is therefore a reasonable baseline: if $f$ is close to linear, we may expect the estimator to behave well. Indeed, there is a theoretical and experimental evidence~\cite{ST} that in typical neural networks the more units are used per layer, the closer we are to the linear regime (at least initially) and the better the utility of the estimate for optimization. Furthermore,~\cite{PSA} show that in SBNs of moderate size, the accuracy of \ST estimator is on par with a more accurate PSA estimator.

We will study alternative single-sample approaches and improvements proposed to the basic \ST. 
In order to analyze BayesBiNN and FouST we will switch to the $\pm1$ encoding.
%
%In \cref{sec:FouST} we will switch to the $\pm1$ encoding. 
We will write $y\sim \Bin(p)$ to denote a random variable with values $\{-1,1\}$ parametrized by $p = \Pr_y(y{=}1)$. Alternatively, we will parametrize the same distribution using the expectation $\mu = 2p-1$ and denote this distribution as $\Bin(\mu)$ (the naming convention and the context should make it unambiguous).
%the expectation $\mu$. 
Note that the mean of $\Bernoulli(p)$ is $p$.
%\begin{proposition}
The \ST estimator of the gradient in the mean parameter $\mu$ in both $\{0,1\}$ and $\{-1,1\}$ valued cases is conveniently given by the same equation~\eqref{ST}.
%\end{proposition}
%If the gradient in the probability $p = \Pr(y{=}1)$ is requested, it expresses as
%\begin{align}
%\frac{\d}{\d p}f = 2 \frac{\d}{\d \mu}f.
%\end{align}
%
%that $y = 1$ with probability $p$ and $-1$ with probability $1-p$. 
%This case is related to the Boolean encoding by the obvious 
%\begin{proposition}
%Let $\hat g$ be a stochastic estimator of gradient $\E_x [f(x)]$ in $p$, where $x\sim \Bernoulli(p)$. Then, under change of variables $y=2x -1$, $\hat g$
%\begin{align}
%\end{align}
%%$\mu = 2p-1$
%%Then $\hat g$ is also the correct gradient estimator $\E_y [\tilde f(y)]$ in $p$, where $x\sim \Bernoulli(p)$
%%$\hat g$ is also the corresponding for $y\sim \Bin(\mu)$ 
%\end{proposition}
\begin{proof}
Indeed, $\E_y [f(y)]$ with $y\sim \Bin(\mu)$ can be equivalently expressed as $\E_x [\tilde f(x)]$ with $x\sim \Bernoulli(p)$, 
where $p = \frac{\mu + 1}{2}$ and $\tilde f(x) = f(2x -1)$. 
%where $x = (y+1)/2$, $p = \frac{\mu + 1}{2}$ .
The \ST estimator of gradient in the Bernoulli probability $p$ for a sample $x$ can then be written as
\begin{align}\label{ST-mu}
\gst = \tilde f'(x) =  2 f'(y),
\end{align}
where $y= 2x-1$ is a sample from $\Bin(\mu)$.
The gradient estimate in $\mu$ becomes $2 f'(y) \frac{\partial p}{\partial \mu} = f'(y)$.
\end{proof}
%and will try to always compare them to
%
%
%The total variance is therefore
%\begin{align}
%\V[\gst ] &= \E_y[\V_x[\gst ]] + \V_y[\E_x[\gst ]] = 4 \V_y[ h(y)].
%\end{align}
%

%%%%%%%%%%%%%%%%%%%%%%%%%%%%%%%%%%%%%%%%%%%%%%%%%%%%%%%%%%%%%%%%
\section{Gumbel Softmax and ST Gumbel-Softmax}\label{sec:GS}

Gumbel Softmax~\citep{jang2016categorical} and Concrete relaxation~\citep{maddison2016concrete} enable differentiability through discrete variables by relaxing them to real-valued variables that follow a distribution closely approximating the original discrete distribution. The two works~\cite{jang2016categorical,maddison2016concrete} have contemporaneously introduced the same relaxation, but the name Gumbel Softmax (GS) became more popular in the literature. 

A categorical discrete random variable $x$ with $K$ category probabilities $\pi_k$ can be sampled as
\begin{align}
x = \argmax_k(\log \pi_k - \Gamma_k ),
\end{align}
where $\Gamma_k$ are independent Gumbel noises. %In this section and in the respective proofs in~\cref{sec:GSA} we use the ${0,1}$ encoding.
This is known as Gumbel reparametrization. In the binary case with categories $k\in\{1,0\}$ we can express it as
\begin{align}
x = \leftbb \log \pi_1 - \Gamma_{1} \geq \log \pi_0 - \Gamma_{0} \rightbb,
\end{align}
where $\leftbb \cdot \rightbb$ is the Iverson bracket. More compactly, denoting $p=\pi_1$,
\begin{align}
x = \leftbb \log \frac{p}{1-p} - (\Gamma_{1} - \Gamma_{0}) \geq 0 \rightbb.
\end{align}
The difference of two Gumbel variables $z = \Gamma_{1} - \Gamma_{0}$ follows the logistic distribution. Its cdf is $\sigma(z) = \frac{1}{1+e^{-z}}$.
Denoting $\eta = \logit(p)$, we obtain the well-known noisy step function representation:
\begin{align}
x = \leftbb \eta - z \geq 0 \rightbb.
\end{align}
%. The issue, commonly known as ``step function being not differentiable'' occurs
This reparametrization of binary variables is exact but does not yet allow for differentiation of a single sample because we cannot take the derivative under the expectation of this function in~\eqref{g-def}. The relaxation~\citep{jang2016categorical,maddison2016concrete} replaces the threshold function by a continuously differentiable approximation $\sigma_\tau(\eta) := \sigma(\eta/\tau) = \frac{1}{1+e^{-\eta/\tau}}$. As the temperature parameter $\tau>0$ decreases towards $0$, the function $\sigma_\tau(\eta)$ approaches the step function. %For a fixed temperature $\tau$ the relaxed sample is obtained as $$. 
%It becomes continuous and differentiable in $\eta$. 
The \GS estimator of the derivative in $\eta$ is then defined as the total derivative of $f$ at a random relaxed sample:
\begin{subequations}\label{GS-grad}
\begin{align}
z & \sim \text{Logistic},\\
\tilde x & = \sigma_\tau(\eta - z),\\
\frac{\hat d f} {d \eta} & := \frac{\d f (\tilde x)}{\d \eta} = f'(\tilde x) \frac{\partial \tilde x}{\partial \eta}.
\end{align}
\end{subequations}

\begin{figure}[t]
\centering
\includegraphics[width=0.4\linewidth]{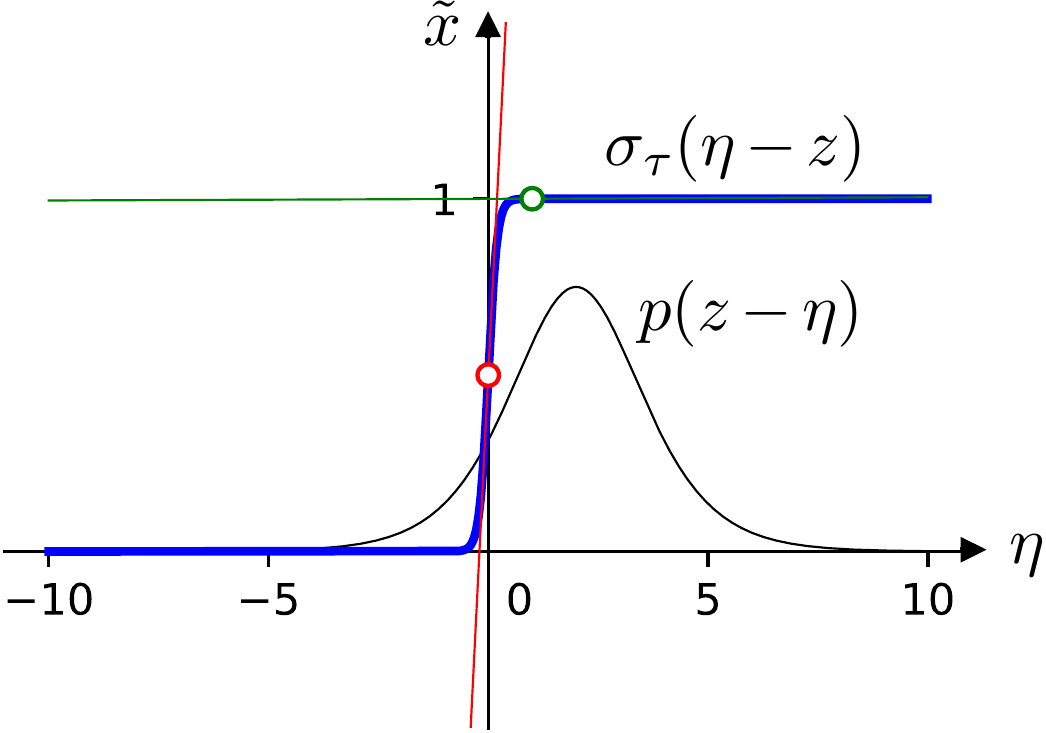}
\caption{GS Estimator: relaxed samples $\tilde x$ are obtained and differentiated as follows. Noisy inputs, following a shifted logistic distribution (black density), are passed through a smoothed step function $\sigma_\tau$ (blue). Observe that for a small $\tau$, the derivative is often (in probability of $\eta-z$) close to zero (green) and, very rarely, when $|\eta-z|$ is small, it becomes $O(1/\tau)$ large (red).
\label{fig:GS}}
\end{figure}

A possible delusion about \GS gradient estimator is that it can be made arbitrary accurate by using a sufficiently small temperature $\tau$. This is however not so simple and we will clarify theoretical reasons for why it is so. An intuitive explanation is proposed in~\cref{fig:GS}. Formally, we show the following properties.

\begin{restatable}{proposition}{PropGS}\label{PropGS1}
\GS estimator is asymptotically unbiased as $\tau \rightarrow 0$ and the bias decreases at the rate $O(\tau)$ in general and at the rate $O(\tau^2)$ for linear functions.
\end{restatable}
\noindent Proof in Appendix A. %First note that the estimator is biased even for linear objectives. 
The decrease of the bias with $\tau\rightarrow 0$ is a desirable property, but this advantage is practically nullified by the fast increase of the variance:

\begin{restatable}{proposition}{PropGSV}\label{PropGSV}
The variance of \GS estimator grows at the rate $O(\frac{1}{\tau})$.
\end{restatable}
\noindent Proof in Appendix A. This fast growth of the variance prohibits the use of small temperatures in practice.
In more detail the behavior of the gradient estimator is described by the following two propositions.

\begin{restatable}{proposition}{PropGSB}\label{PropGS2}
For any given realization $z \neq \eta$ the norm of \GS estimator asymptotically vanishes at the exponential rate $O(\frac{1}{\tau} c^{1/\tau})$ with $c=e^{-|x|} < 1$.
\end{restatable}
\noindent Proof in Appendix A. For small $x$, where $c$ is close to one, the term $1/\tau$ dominates at first. In particular for $z=\eta$, the asymptote is $O(1/\tau)$. So while for the most of noise realizations the gradient magnitude vanishes exponentially quickly, it is compensated by a significant grows at rate $1/\tau$ around $z=\eta$. In practice it means that most of the time a value of gradient close to zero is measured and occasionally, very rarely, a value of $O(1/\tau)$ is obtained.

\begin{restatable}{proposition}{PropGSC}\label{PropGS3}
The probability to observe \GS gradient of norm at least $\varepsilon$ is asymptotically $O(\tau \log(\frac{1}{\varepsilon}))$, where the asymptote is $\tau \rightarrow 0$, $\varepsilon \rightarrow 0$.
\end{restatable}
\noindent Proof in Appendix A.

%Since the gradient is asymptotically unbiased, such quick diminishing for any fixed $\varepsilon$ has to be compensated by a rapid growth around $\eta$.

Unlike \ST, \GS estimator with $\tau>0$ is biased even for linear objectives. Even for a single neuron and a linear objective it has a non-zero variance.
\cref{PropGS2,PropGS3} apply also to the case of a layer with multiple units since they just analyze the factor $\frac{\partial}{\partial \eta}\sigma_\tau(\eta -z)$, which is present independently at all units. \cref{PropGS3} can be extended to deep networks with L layers of Bernoulli variables, in which case the chain derivative will encounter $L$ such factors and we obtain that the probability to observe a gradient with norm at least $\varepsilon$ will vanish at the rate $O(\tau^L)$.

%The proofs are given below. Basically all 
These facts should convince the reader of the following: it is not possible to use a very small $\tau$, not even with an annealing schedule starting from $\tau=1$. For a very small $\tau$ the most likely consequence would be to never encounter a numerically non-zero gradient during the whole training. For moderately small $\tau$ the variance would be prohibitively high. Indeed,~\citet{jang2016categorical} anneal $\tau$ only down to $0.5$ in their experiments.

%\revisit[Issue with relaxed samples through deep network]
%Furthermore, the following issue occurs when the estimator is used in deep networks. The forward pass uses relaxed samples $\tilde x$, which biases all the expectations for all other units in the same or deeper layers. 

A major issue with this and other relaxation techniques (\ie techniques using relaxed samples $\tilde x\in \Real$) is that the relaxation biases all the expectations. There is only one forward pass and hence the relaxed samples $\tilde x$ are used for all purposes, not only for the purpose of estimating the gradient with respect to the given neuron. It biases all expectations for all other units in the same layer as well as in preceding and subsequent layers (in SBN).
%
%$f$ is evaluated at a relaxed points and this value is used for all purposes, not only in the aforementioned gradient. It affects all expectations as $E_{\tilde x}[f(\tilde x)]$ is generally different from $E_{x}[f(x)]$. 
%
Let for example $f$ depend on additional parameters $\theta$ in a differentiable way. More concretely, $\theta$ could be parameters of the decoder in VAE. With a Bernoulli sample $x$, an unbiased estimate of gradient in $\theta$ can be obtained simply as $\frac{\partial}{\partial \theta }f(x;\theta)$. However, if we replace the sample with a relaxed sample $\tilde x$, the estimate $\frac{\partial}{\partial \theta }f(\tilde x;\theta)$ becomes biased because the distribution of $\tilde x$ only approximates the distribution of $x$.
%
%
% and thus $\frac{\partial}{\partial \theta}E_{x,u}[f(zu;\theta)]$ will be a biased estimate of $\frac{\partial}{\partial \theta}E_{x}[f(x;\theta)]$. 
%
If $y$ were other binary variables relaxed in a similar way, the gradient estimate for $x$ will become more biased because $E_{\tilde y}[\nabla_{\tilde x}f(\tilde x, \tilde y)]$ is a biased estimate of $E_{y}[\nabla_{\tilde x}f(\tilde x, y)]$ desired. Similarly, in a deep SBN, the relaxation applied in one layer of the model additionally biases all expectations for units in layers below and above. In practice the accuracy for VAEs is relatively good~\cite{jang2016categorical},~\cite[Fig. 3]{ST} while for deep SBNs a bias higher than of ST is observed for $\tau=1$ in a synthetic model with 2 or more layers and with $\tau=0.1$ for a model with 7 (or more) layers~\cite[Fig. C.6]{PSA}. When training moderate size SBNs on real data, it performs worse than ST~~\cite[Fig. 4]{PSA}.

\paragraph{ST Gumbel-Softmax}
Addressing the issue that relaxed variables deviate from binary samples on the forward pass,~\citet{jang2016categorical} proposed the following empirical modification. {\em ST Gumbel-Softmax} estimator~\cite{jang2016categorical} keeps the relaxed sample for the gradient but uses the correct Bernoulli sample on the forward pass:
\begin{subequations}
\begin{align}
z & \sim \text{Logistic},\\
\tilde x & = \sigma_\tau(\eta - z),\\
x & = \leftbb \eta - z \geq 0 \rightbb,\\
\hat g_{\text{\sc st-gs}(\tau)} & = f'(x) \frac{\partial \tilde x}{\partial \eta}.
\end{align}
\end{subequations}

%\begin{align}
%\frac{\partial \L(x_1)}{\partial x_1} \frac{\partial y_1}{\partial \eta},
%\end{align}
%where both $x_1$ and $y_1$ are computed using the same logistic noise $z$. 
\noindent Note that $x$ is now distributed as $\Bernoulli(p)$ with $p =\sigmoid(\eta)$ so the forward pass is fixed. We show the following asymptotic properties.
%However, the gradient estimate is worsened:
\begin{restatable}{proposition}{PropSTGS}\label{PropSTGS}
%The ST Gumbel-softmax estimator~\citep{jang2016categorical} is asymptotically (for  $\tau\rightarrow 0$) unbiased for quadratic functions  even for a linear loss.
ST Gumbel-Softmax estimator~\citep{jang2016categorical} is asymptotically unbiased for quadratic functions and the variance grows as $O(1/\tau)$ for $\tau \rightarrow 0$. %is asymptotically (for  $\tau\rightarrow 0$) unbiased for quadratic functions  even for a linear loss.
\end{restatable}
\noindent Proof in Appendix A.

%\revisit
To summarize, \ST-\GS is more expensive than \ST as it involves sampling from logistic distribution (and keeping samples), it is biased for $\tau>0$. It becomes unbiased for quadratic functions as $\tau \rightarrow 0$, which would be an improvement over ST, but the variance grows as $\frac{1}{\tau}$.

%%%%%%%%%%%%%%%%%%%%%%%%%%%%%%%%%%%%%%%%%%%%%%%%%%%%%%%%%%%%%%%%%%%%%%%%%%%%%%%%%%%%%%%%%%%%%%%%%
\section{BayesBiNN}\label{sec:BayesBiNN}
\citet{meng-20}, motivated by the need to reduce the variance of \REINFORCE, apply GS estimator. However, in their large-scale experiments they use temperature $\tau = 10^{-10}$. According to the previous section, the variance of \GS estimator should go through the roof as it grows as $O(\frac{1}{\tau})$. It is practically prohibitive as the learning would require an extremely small learning rate and a very long training time as well as high numerical accuracy. Nevertheless, good experimental results are demonstrated~\cite{meng-20}. We identify a hidden implementation issue which completely changes the gradient estimator and enables learning.

First, we explain the issue. \citet{meng-20} model stochastic binary weights as $w\sim \Bin(\mu)$ and express \GS estimator as follows.
\begin{restatable}[\citet{meng-20} Lemma 1]{proposition}{PropGStanh}\label{PropGStanh}
Let $w\sim \Bin(\mu)$ and let $f\colon \{-1,1\}\to \Real$ be a loss function.
Using parametrization $\mu = \tanh(\lambda)$, $\lambda\in\Real$, \GS estimator of gradient $\frac{\d \E_w[f]}{\d \mu}$ can be expressed as
\begin{subequations}
\begin{align}
\delta & \sim \frac{1}{2}\text{\rm Logistic},\\
\tilde w & = \tanh_{\tau}(\lambda - \delta) \equiv \tanh(\frac{\lambda - \delta}{\tau}),\\
J &= \frac{1 - \tilde w^2}{\tau(1 - \mu^2)},\\
\hat g & = J f'(\tilde w),
%\ggst & = f'(\tilde w) \frac{\partial \tilde w}{\partial \lambda}.
\end{align}
\end{subequations}
%where $J = (1 - \tilde w^2)/(\tau(1 - \mu^2))$.
\end{restatable}
\noindent which we verify in Appendix B.
However, the actual implementation of the scaling factor $J$ %$J = \frac{1 - \tilde w^2}{\tau(1 - \mu^2)}$ 
used in the experiments~\cite{meng-20} according to the published code\footnote{\url{https://github.com/team-approx-bayes/BayesBiNN}} introduces a technical $\epsilon=10^{-10}$ as follows: 
\begin{align}\label{BayesBiNN-J}
J := \frac{1-\tilde w^2 + \epsilon}{\tau (1-\mu^2 + \epsilon)}.
\end{align}
It turns out this changes the nature of the gradient estimator and of the learning algorithm.
%We will show that with this technical $\varepsilon$, the estimator becomes a deterministic straight-through estimator in the learning algorithm.
The BayesBiNN algorithm~\citep[][Table 1 middle]{meng-20} performs the update:
\begin{align}\label{BBN-lambda-step}
\lambda  &:= (1-\alpha) \lambda - \alpha s f'(\tilde w),
\end{align}
where $s = N J$, $N$ is the number of training samples and $\alpha$ is the learning rate.  
%according to~\cite[Eq. 9]{meng-20},: %a scaling factor combining the Jacobian $\frac{\partial w}{\partial p}$ from Gumbel-Softmax estimator and the evidence-to-prior scaling N from the VB formulation~\cite[Eq. 9]{meng-20}:
%\begin{align}\label{BayesBiNN}
%s = N \frac{1-w^2}{\tau (1-\tanh(\eta)^2)}.
%\end{align}
%It can be verified that $\alpha s g/\varepsilon$ equals $f'(w) \frac{\partial w}{\partial \mu}$, the \GS gradient estimator.
%The actual implementation of the scaling factor $s$ used in the experiments~\cite{meng-20} according to the published code~\footnote{\url{https://github.com/team-approx-bayes/BayesBiNN}} introduces a technical $\varepsilon=10^{-10}$: 
%\begin{align}\label{BayesBiNN}
%s = N \frac{1-w^2 + \varepsilon}{\tau (1-\tanh(\eta)^2 + \varepsilon)}.
%\end{align}

\begin{restatable}{proposition}{PropBayesBiNNA}\label{PropBayesBiNNA}
With the setting of the hyper-parameters $\tau = O(10^{-10})$~\citep[][Table 7]{meng-20} and $\epsilon = 10^{-10}$~(author's implementation) in large-scale experiments (MNIST, CIFAR10, CIFAR100), the BayesBiNN algorithm is practically equivalent to %SGD with deterministic identity straight-through and latent weight decay.
%SMD using a deterministic \ST and latent weight decay: % that does not depend on the data set size $N$.
the following deterministic algorithm:
\begin{subequations}\label{BayesBiNN-step-R1}
\begin{align}
w &:= \sign(\bar \lambda);\\
\bar \lambda &:= (1-\alpha) \bar\lambda - \alpha f'(w).\label{BayesBiNN-step-R1-b}
\end{align}
\end{subequations}
In particular, it does not depend on the values of $\tau$ and $N$.
\end{restatable}
\noindent Proof in Appendix B. Experimentally, we have verified, using authors implementation, that indeed parameters $\lambda$~\eqref{BBN-lambda-step} grow to the order $10^{10}$ during the first iterations, as predicted by our calculations in the proof.

Notice that the step made in~\eqref{BayesBiNN-step-R1-b} consists of a decay term $-\alpha \bar\lambda$ and the gradient descent term $- \alpha f'(w)$, where the gradient $f'(w)$ is a straight-through estimate for the {\em deterministic} forward pass $w = \sign(\bar \lambda)$. Therefore the deterministic \ST is effectively used. It is seen that the decay term is the only remaining difference to the deterministic STE algorithm~\citep[][Table 1 left]{meng-20}, the method is contrasted to. From the point of view of our study, we should remark that the deterministic ST estimator used in effect indeed decreases the variance (down to zero) however it increases the bias compared to the baseline stochastic \ST~\cite{ST}.

The issue has also downstream consequences for the intended Bayesian learning.
The claim of~\cref{PropBayesBiNNA} that the method does not depend on $\tau$ and $N$ is perhaps somewhat unexpected, but it makes sense indeed. The initial BayesBiNN algorithm of course depends on $\tau$ and $N$.
However due to the issue with the implementation of Gumbel Softmax estimator, for a sufficiently small value of $\tau$ it falls into a regime which is significantly different from the initial Bayesian learning rule and is instead more accurately described by~\eqref{BayesBiNN-step-R1}. 
In this regime, the result it produces does not dependent on the particular values of $\tau$ and $N$. While we do not know what problem it is solving in the end, it is certainly not solving the intended variational Bayesian learning problem.  This is so because the variational Bayesian learning problem and its solution do depend on $N$ in a critical way. The algorithm~\cref{BayesBiNN-step-R1} indeed does not solve any variational problem as there is no variational distribution involved (nothing sampled). Yet, the decay term $-\alpha \lambda$ stays effective: if the data gradient becomes small, the decay term implements some small ``forgetting'' of the learned information and may be responsible for an improved generalization observed in the experiments~\citep{meng-20}. 
%

%%%%%%%%%%%%%%%%%%%%%%%%%%%%%%%%%%%%%%%%%%%%%%%%%%%%%%%%%%%%%%%%%%%%%%%%%%%%%%%%%%%%%%%%%%%%%%%%%

\section{FouST}\label{sec:FouST}
\citet{Pervez-20} introduced several methods to improve ST estimators using Fourier analyzes of Boolean functions~\cite{Donnell-14} and Taylor series. The proposed methods are guided by this analysis but lack formal guarantees. We study the effect of the proposed improvements analytically. 

One issue with the experimental evaluation~\cite{Pervez-20} is that the baseline ST estimator~\cite[Eq. 7]{Pervez-20} is {\em misspecified}: it is adopted from the works considering $\{0,1\}$ Bernoulli variables without correcting for $\{-1,1\}$ case as in~\eqref{ST-mu}, differing by a coefficient $2$. The reason for this misspecifications is that ST is know rather as a folklore, vaguely defined, method (see~\cite{ST}).
While in learning with a simple expected loss this coefficient can be compensated by the tuned learning rate, it can lead to a more serious issues, in particular in VAEs with Bernoulli latents and deep SBNs. VAE training objective~\cite{KingmaW13} has the data evidence part, where binary gradient estimator is required and the prior KL divergence part, which is typically computed analytically and differentiated exactly. Rescaling the gradient of the evidence part only introduces a bias which cannot be compensated by tuning the learning rate. Indeed, it is equivalent to optimizing the objective with the evidence part rescaled. In~\cite[Fig. 2]{ST} we show that this effect is significant. 
In the reminder of the section we will assume that the correct ST estimator~\eqref{ST-mu} is used as the starting point.

%The are demonstrated to give a significant improvement over ST in several tasks. The implementation to reproduces these experiments is however not published.
\subsection{Lowering Bias by Importance Sampling}
The method~\cite[Sec. 4.1]{Pervez-20} "Lowering Bias by Importance Sampling", as noted by authors, obtains \DARN gradient estimator~\cite[Appendix A]{Gregor-14} who derived it by applying a (biased) control variate estimate in the \REINFORCE method. %This estimator is denoted as ``$\frac{1}{2}$'' estimator in~\cite[Section 4.2]{MuProp}, where experimental evaluation. 
Transformed to the encoding with $\pm 1$ variables, it expresses as
\begin{align}\label{DARN}
\gdarn  = f'(x)/p(x).
\end{align}
By design~\cite{Gregor-14}, this method is unbiased for quadratic functions, which is straightforward to verify by inspecting its expectation
\begin{align}\label{E-DARN}
\E[\gdarn]  = f'(1) + f'(0).
\end{align}
While, this is in general an improvement over ST --- we may expect that functions close to linear will have a lower bias, it is not difficult to construct an example when it can increase the bias compared to ST.
\begin{example}
The method~\cite[Sec. 4.1]{Pervez-20} "Lowering Bias by Importance Sampling", also denoted as Importance Reweighing (IR), can increase bias.

Let $p\in[0,1]$ and $x \sim \Bin(p)$. Let $f(x) = |x + a|$. The derivative of $\E[f(x)]$ in $p$ is 
\begin{align}
\frac{\d}{\d p} ((1-p) f(-1) + p f(1)) = f(1) - f(-1) = f(1) = 2a.
\end{align}
The expectation of $\gst $ is given by
\begin{align}
(1-p) 2 f'(-1) + p 2 f'(1)  = 2 (2p-1).
\end{align}
The expectation of $\gdarn$ is given by
\begin{align}
f'(-1) + f'(1)  = 0.
\end{align}
The bias of \DARN is $2|a|$ while the bias of ST is $2|a+1-2p|$. Therefore for $a>0$ and $p>0.5$, the bias of \DARN estimator is higher. In particular for $a=0.9$ and $p=0.95$ the bias of ST estimator equals $0$ while the bias of \DARN estimator equals $1.8$. 
\end{example}
Furthermore, we can straightforwardly express its variance.
\begin{proposition}
The variance of $\gdarn$ is expressed as
\begin{align}
\V_z [\hat g_{\text{\sc darn}}] = \frac{(f'(1) - p (f'(1) + f'(-1)))^2}{p(1-p)}.
\end{align}
It has asymptotes $O(\frac{f'(-1)^2}{1-p})$ for $p\rightarrow 1$ and $O(\frac{f'(1)^2}{p})$ for $p\rightarrow 0$. %The variance does not vanish even for linear functions unless $p=\frac{1}{2}$.
\end{proposition}

%There exist cases indeed where \DARN estimator gives a lower bias {\em and} a lower variance. In particular, if $p = \frac{f'(-1)}{f'(-1) + f'(1)}$, the variance vanishes. 
The asymptotes indicate that the variance can grow unbounded for units approaching deterministic mode. If applied in a deep network with $L$ layers, $L$ expressions~\eqref{DARN} are multiplied  and the variance can grow respectively. Interestingly though, if the probability $p$ is defined using the sigmoid function as $p=\sigma(\eta)$, then the gradient in $\eta$ additionally multiplies by the Jacobian $\sigma'(\eta) = p (1-p)$, and the variance of the gradient in $\eta$ becomes bounded. Moreover, a numerically stable implementation can simplify $p(1-p)/p(x)$ for both outcomes of $x$. We conjecture that this estimator can be particularly useful with this parametrization of the probability (which is commonly used in VAEs and SBNs).

\begin{figure}[t]
\setlength{\tabcolsep}{0pt}
\begin{tabular}{ccc}
\includegraphics[width=0.34\linewidth]{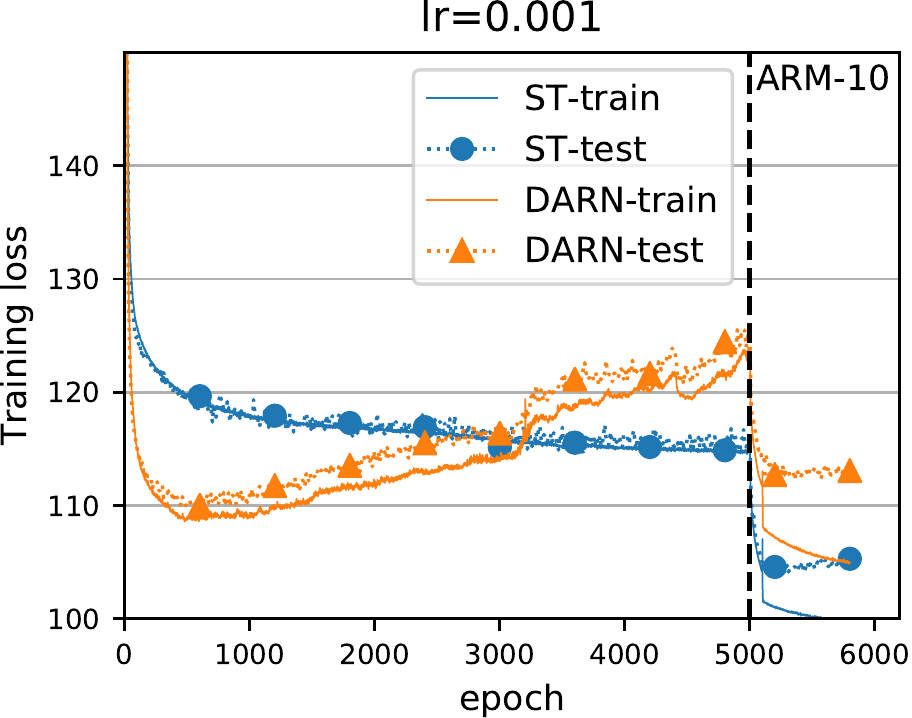}&
\includegraphics[width=0.32\linewidth]{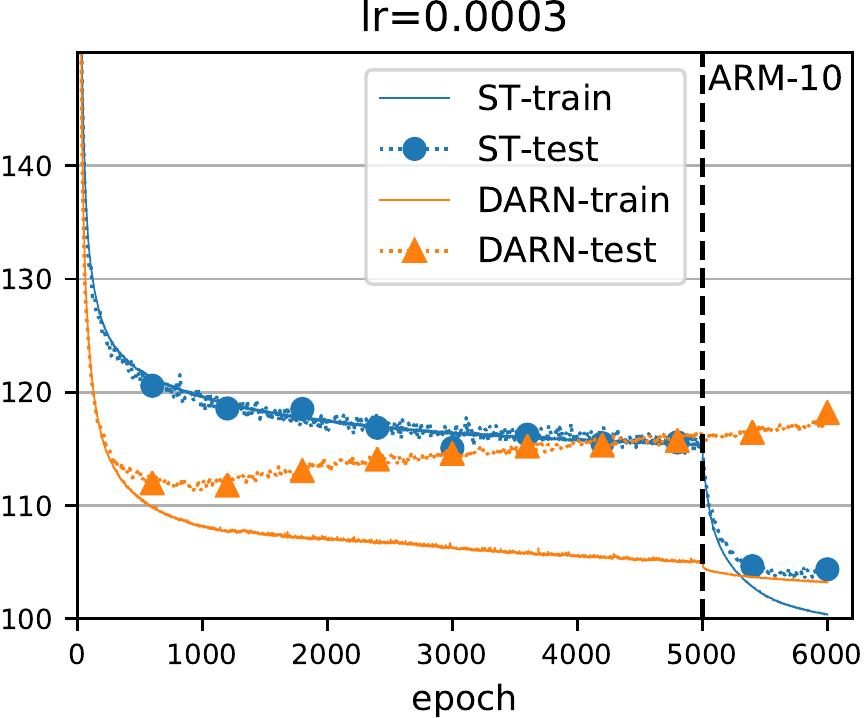}&
\includegraphics[width=0.32\linewidth]{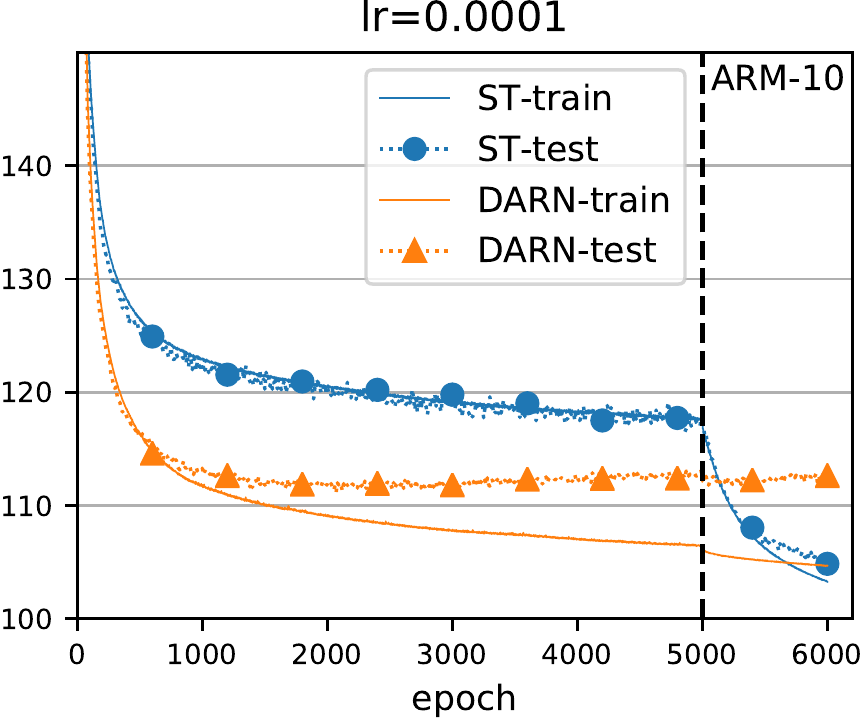}
\end{tabular}
\caption{Experimental comparison of \DARN and \ST estimators on MNIST VAE. The plots show training and test loss (negative ELBO) during training for different learning rates. After 5000 epochs, an unbiased ARM-10 estimator is applied in order to measure (and correct) the accumulated bias. At the smaller learning rates, where DARN does not diverge, it clearly has a much smaller accumulated bias but manages to overfit significantly.
%\DARN appears to give improvement in the beginning, where, due to random initialization, probabilities of hidden states are closer to $1/2$. As training progresses, it starts suffering from higher variance as visible at larger learning rates. It appears to give a marginal improvement over ST for a small learning rate, but the absolute achieved loss value is higher than that for ST with a larger learning rate.
\label{fig:darn}}
\end{figure}

Experimental evidence~\cite[Fig 2.a]{MuProp}, where \DARN estimator is denoted as ``$\frac{1}{2}$'' shows that the plain \ST performs similar for the structural output prediction problem. However,~\cite[Fig 3.a]{MuProp} gives a stronger evidence in favor of DARN for VAE. In~\cref{fig:darn} we show experiment for the MNIST VAE problem, reproducing the experiment~\cite{MuProp,Pervez-20} (up to data binarization and implementation details). The exact specification is given in~\cite[Appendix D.1]{ST}. It is seen that \DARN improves the training performance compared but needs an earlier stopping and or more regularization. Interestingly, with a correction of accumulated bias using unbiased ARM~\cite{yin18-arm} method with 10 samples, ST leads to better final training and test performance. 
%In our experiment \ST still performed better for this task.

%\begin{proposition} For linear functions, variance of DARN estimator is never lower than variance of ST estimator.
%\end{proposition}

\subsection{Reducing Variance via the Fourier Noise Operator}
The Fourier noise operator~\cite[Sec. 2]{Pervez-20} is defined as follows.
For $\rho \in[0,1]$, let $x'\sim N_\rho(x)$ denote that $x'$ is set equal to $x$ with probability $\rho$ and chosen as an independent sample from $\Bin(p)$ with probability $1-\rho$.
The Fourier noise operator smooths the loss function and is defined as $T_\rho[f](x) = \E_{x'\sim N_\rho(x)}[f(x')]$. When applied to $f$ before taking the gradient, it can indeed reduce both bias and variance, ultimately down to zero when $\rho=0$. Indeed, in this case $x'$ is independent of $x$ and $T_\rho[f](x) = \E[f(x)]$, which is a constant function of $x$. However, the exact expectation in $x'$ is intractable. The computational method proposed in~\cite[Sec. 4.2]{Pervez-20} approximates the gradient of this expectation using $S$ samples $x^{(s)}\sim N_\rho(x)$ as %. The estimator that is proposed is defined as
\begin{align}
\hat g_{\rho} = \frac{1}{S} \sum_{s} \hat g (x^{(s)}),
\end{align}
where $\hat g$ is the base \ST or \DARN estimator. We show the following.

\begin{restatable}{proposition}{PropFouSTNosie}\label{PropFouSTNosie}
The method~\cite[Sec. 4.2]{Pervez-20} "Reducing Variance via the Fourier Noise operator" does not reduce the bias (unlike $T_\rho$) and increases variance in comparison to the trivial baseline that averages independent samples.
\end{restatable}
\noindent Proof in~Appendix C.

This result is found in a sharp contradiction with the experiments~\cite[Figure 4]{Pervez-20}, where independent samples perform worse than correlated. We do not have a satisfactory explanation for this discrepancy except for the misspecified \ST. Since the author's implementation is not public, it is infeasible to reproduce this experiment in order to verify whether a similar improvement can be observed with the well-specified \ST. Lastly, note, that unlike correlated sampling, uncorrelated sampling can be naturally applied with multiple stochastic layers.
%
%\revisit[Cannot be applied in multiple stochastic layers]
%
%
\subsection{Lowering Bias by Discounting Taylor Coefficients}
For the technique~\cite[Sec. 4.3.1]{Pervez-20} "Lowering Bias by Discounting Taylor Coefficients" we present an alternative view, not requiring Taylor series expansion of $f$, thus simplifying the construction. Following~\cite[Sec. 4.3.1]{Pervez-20} we assume that the importance reweighing was applied. Since the technique samples $f'$ at non-binary points, we refer to it as a {\em relaxed} \DARN estimator. It can be defined as 
\begin{align}\label{relaxed-darn}
\tilde g_{\text{\sc darn}}(x,u) = \frac{f'(x u)}{p(x)}, \text{where \ } u\sim \U[0,1].
\end{align}
In the total expectation, when we draw $x$ and $u$ multiple times, the gradient estimates are averaged out. The expectation over $u$ alone effectively integrates the derivative to obtain:
\begin{align}
\E_u \big[\tilde g_{\text{\sc darn}}(x,u)] = 
\begin{cases}
\frac{1}{p}\int_{0}^{1} f'(u) \d u = \frac{1}{p}(f(1) - f(0)), & \text{if\ \ } x = 1,\\
\frac{1}{1-p}\int_{0}^{1} f'(-u) \d u = \frac{1}{1-p}(f(-1) - f(0)), & \text{if \ } x = -1.\\
\end{cases}
%\frac{f'(x u)}{p(x)}, \text{where} u\sim \U[0,1].
\end{align}
In the expectation over $x$ we therefore obtain
\begin{align}\label{relaxed-darn-E}
\E_{x,u} [\tilde g_{\text{\sc darn}}(x,u)] = f(1) - f(-1),
\end{align}
which is the correct derivative. One issue, discussed by~\cite{Pervez-20} is that variance increases (as there is more noise in the system).
However, a major issue similar to \GS estimator~\cref{sec:GS}, reoccurs here, that all related expectations become biased. In particular~\eqref{relaxed-darn} becomes biased in the presence of other variables. \citet[Sec. 4.3.1]{Pervez-20} propose to use $u\in\U[a, 1]$ with $a>0$, corresponding to shorter integration intervals around $\pm 1$ states, in order to find an optimal tradeoff. %This gives up the property~\eqref{relaxed-darn} in favor of a smaller increase of the bias in the simultaneous expectations.

%However, a major issue with this technique is that $f$ is evaluated at a relaxed points and this value is used for all purposes, not only in the aforementioned gradient. It affects all expectations as $E_{\tilde x}[f(\tilde x)]$ is generally different from $E_{x}[f(x)]$. Let for example $f$ depend on additional parameters $\theta$ and differentiable in $\theta$. The gradient in $\theta$ can be estimated simply as $\frac{\partial}{\partial \theta }f(x;\theta)$. However, if we replace $x$ with a relaxed sample $\tilde x = xu$, the estimate $\frac{\partial}{\partial \theta }f(xu;\theta)$ becomes biased.
%%
%% and thus $\frac{\partial}{\partial \theta}E_{x,u}[f(zu;\theta)]$ will be a biased estimate of $\frac{\partial}{\partial \theta}E_{x}[f(x;\theta)]$. 
%For example, in the VAE setting, it will lead to a biased estimate of the gradient in decoder parameters.
%If $y$ were other binary variables relaxed to $\tilde y$ in a similar way, the gradient estimate in $p$ for $x$ will also become biased because
%$E_{\tilde y}[f(1,\tilde y) - f(-1,\tilde y)]$ will be a biased estimate of $E_{y}[f(1,y) - f(-1,y)]$.
%
%
%
%\begin{proposition}
%The method~\cite[Sec. 4.3.1]{Pervez-20} "Lowering Bias by Discounting Taylor Coefficients" can increase bias.
%\end{proposition}
\subsection{Lowering Bias by Representation Rescaling}
Consider the estimator $\hat g$ of the gradient of function $\E_x[f(x)]$ where $x \sim \Bin(p)$.
Representation rescaling is defined in~\cite[Algorithm 1]{Pervez-20} as drawing $\tilde x \sim \frac{1}{\tau} \Bin(p)$ instead of $x$ and then using FouST estimator based on the derivative $f'(\tilde x)$. It is claimed that using a scaled representation can decrease the bias of the gradient estimate. However, the following issue occurs.
%The gradient estimate is then based on $f(\tilde x)$.
\begin{proposition}
The method~\cite[Sec. 4.3.2]{Pervez-20} "Lowering Bias by Representation Rescaling" compares biases of gradient estimators of different functions. 
\end{proposition}
\begin{proof}
 Sampling $\tilde x$ can be equivalently defined as $\tilde x = x/\tau$. %Clearly, the two sampling schemes differ  $x$ and set $\tilde x = x/\tau$.
Bypassing the analysis of Taylor coefficients~\cite{Pervez-20}, it is easy to see that for a smooth function $f$, as $\tau \rightarrow \infty$, $f(x/\tau)$ approaches a linear function of $x$ and therefore the bias of the \ST  estimator of $\E_x[f(x/\tau)]$ approaches zero. However, clearly $\E_x[f(x/\tau)]$ is a different function from $\E_x[f(x)]$ which we wish to optimize.
\end{proof}
We explain, why this method nevertheless has effect. Choosing and fixing the scaling hyper-parameter $\tau$ is equivalent to staring from a different initial point, where (initially random) weights are scaled by $1/\tau$. At this initial point, the network is found to be closer to a linear regime, where the ST estimator is more accurate and possibly the vanishing gradient issue is mitigated. Thus the method can have a positive effect on the learning as observed in~\cite[Appendix Table 3]{Pervez-20}.
%Since the initial neural network weights are random, setting $\tau$ has the effect 

\section{Conclusion}\label{sec:conclusion}
We theoretically analyzed properties of several methods for estimation of binary gradients and gained interesting new insights.
\begin{itemize}
\item For GS and ST-GS estimator we proposed a simplified presentation for the binary case and explained detrimental effects of low and high temperatures. We showed that bias of ST-GS estimator approaches that of DARN, connecting these two techniques.
\item For BayesBiNN we identified a hidden issue that completely changes the behavior of the method from the intended variational Bayesian learning with Gumbel-Softmax estimator, theoretically impossible due to the used temperature $\tau=10^{-10}$, to non-Bayesian learning with deterministic ST estimator and latent weight decay. As this learning method shows improved experimental results, it becomes an open problem to clearly understand and advance the mechanism which facilitates this.
\item In our analysis of techniques comprising FouST estimator, we provided additional insights and showed that some of these techniques are not well justified. % and may not lead to improvements, depending on the case in practice. %Our independent reimplementation of the importance reweighing (\DARN) technique showed only a slight improvement in the beginning of the training or when a very small rate is chosen. Since the author's implementation of FouST is not public, we found it infeasible to reproduce their (other) successful experiments. 
It remains open, whether they are nevertheless efficient in practice in some cases for other unknown reasons, not taken into account in this analysis. %despite the lack of %the negative theoretical results we presented.
\end{itemize}
%We cleared the ground for future research.
Overall we believe our analysis clarifies the surveyed methods and uncovers several issues which limit their applicability in practice. 
It provides tools and clears the ground for any future research which may propose new improvements and would need to compare with existing methods both theoretically and experimentally. We hope that this study will additionally motivate such research. %We hope this study will help to understand possible training / convergence problems when applying  and motivate the search for better solutions.
%
%
%\input{tex/Gumbel.tex}
%\input{tex/BayesBiNN.tex}
%
%\clearpage

%\addcontentsline{toc}{section}{References}
{
\small
\bibliographystyle{icml2021}
\bibliography{strings,all,our}

\begin{thebibliography}{36}
\providecommand{\natexlab}[1]{#1}
\providecommand{\url}[1]{\texttt{#1}}
\expandafter\ifx\csname urlstyle\endcsname\relax
  \providecommand{\doi}[1]{doi: #1}\else
  \providecommand{\doi}{doi: \begingroup \urlstyle{rm}\Url}\fi

\bibitem[Alizadeh et~al.(2019)Alizadeh, Fernandez-Marques, Lane, and
  Gal]{alizadeh2018a}
Alizadeh, M., Fernandez-Marques, J., Lane, N.~D., and Gal, Y.
\newblock An empirical study of binary neural networks' optimisation.
\newblock In \emph{ICLR}, 2019.

\bibitem[Bethge et~al.(2019)Bethge, Yang, Bornstein, and Meinel]{Bethge-19}
Bethge, J., Yang, H., Bornstein, M., and Meinel, C.
\newblock Back to simplicity: How to train accurate {BNN}s from scratch?
\newblock \emph{CoRR}, abs/1906.08637, 2019.

\bibitem[Bulat \& Tzimiropoulos(2017)Bulat and Tzimiropoulos]{Bulat_2017_ICCV}
Bulat, A. and Tzimiropoulos, G.
\newblock Binarized convolutional landmark localizers for human pose estimation
  and face alignment with limited resources.
\newblock In \emph{ICCV}, Oct 2017.

\bibitem[Bulat et~al.(2019)Bulat, Tzimiropoulos, Kossaifi, and
  Pantic]{bulat2019improved}
Bulat, A., Tzimiropoulos, G., Kossaifi, J., and Pantic, M.
\newblock Improved training of binary networks for human pose estimation and
  image recognition.
\newblock \emph{arXiv}, 2019.

\bibitem[Bulat et~al.(2021)Bulat, Martinez, and Tzimiropoulos]{bulat2021}
Bulat, A., Martinez, B., and Tzimiropoulos, G.
\newblock High-capacity expert binary networks.
\newblock In \emph{ICLR}, 2021.

\bibitem[Chaidaroon \& Fang(2017)Chaidaroon and Fang]{Chaidaroon-17}
Chaidaroon, S. and Fang, Y.
\newblock Variational deep semantic hashing for text documents.
\newblock In \emph{SIGIR Conference on Research and Development in Information
  Retrieval}, pp.\  75--84, 2017.

\bibitem[Dadaneh et~al.(2020)Dadaneh, Boluki, Yin, Zhou, and
  Qian]{Dadaneh2020PairwiseSH}
Dadaneh, S.~Z., Boluki, S., Yin, M., Zhou, M., and Qian, X.
\newblock Pairwise supervised hashing with {Bernoulli} variational auto-encoder
  and self-control gradient estimator.
\newblock \emph{ArXiv}, abs/2005.10477, 2020.

\bibitem[Esser et~al.(2016)Esser, Merolla, Arthur, Cassidy, Appuswamy,
  Andreopoulos, Berg, McKinstry, Melano, Barch, di~Nolfo, Datta, Amir, Taba,
  Flickner, and Modha]{Esser-16}
Esser, S.~K., Merolla, P.~A., Arthur, J.~V., Cassidy, A.~S., Appuswamy, R.,
  Andreopoulos, A., Berg, D.~J., McKinstry, J.~L., Melano, T., Barch, D.~R.,
  di~Nolfo, C., Datta, P., Amir, A., Taba, B., Flickner, M.~D., and Modha,
  D.~S.
\newblock Convolutional networks for fast, energy-efficient neuromorphic
  computing.
\newblock \emph{Proceedings of the National Academy of Sciences}, 113\penalty0
  (41):\penalty0 11441--11446, 2016.

\bibitem[Grathwohl et~al.(2018)Grathwohl, Choi, Wu, Roeder, and
  Duvenaud]{grathwohl18-relax}
Grathwohl, W., Choi, D., Wu, Y., Roeder, G., and Duvenaud, D.
\newblock Backpropagation through the void: Optimizing control variates for
  black-box gradient estimation.
\newblock In \emph{ICLR}, 2018.

\bibitem[Gregor et~al.(2014)Gregor, Danihelka, Mnih, Blundell, and
  Wierstra]{Gregor-14}
Gregor, K., Danihelka, I., Mnih, A., Blundell, C., and Wierstra, D.
\newblock Deep autoregressive networks.
\newblock In \emph{ICML}, 2014.

\bibitem[Gu et~al.(2016)Gu, Levine, Sutskever, and Mnih]{MuProp}
Gu, S., Levine, S., Sutskever, I., and Mnih, A.
\newblock Muprop: Unbiased backpropagation for stochastic neural networks.
\newblock In \emph{4th International Conference on Learning Representations
  (ICLR)}, May 2016.

\bibitem[{Horowitz}(2014)]{Horowitz-14}
{Horowitz}, M.
\newblock Computing's energy problem (and what we can do about it).
\newblock In \emph{International Solid-State Circuits Conference Digest of
  Technical Papers (ISSCC)}, pp.\  10--14, 2014.

\bibitem[Jang et~al.(2017)Jang, Gu, and Poole]{jang2016categorical}
Jang, E., Gu, S., and Poole, B.
\newblock Categorical reparameterization with gumbel-softmax.
\newblock In \emph{ICLR}, 2017.

\bibitem[Kingma \& Welling(2013)Kingma and Welling]{KingmaW13}
Kingma, D.~P. and Welling, M.
\newblock Auto-encoding variational {Bayes}.
\newblock \emph{CoRR}, abs/1312.6114, 2013.

\bibitem[Liu et~al.(2018)Liu, Wu, Luo, Yang, Liu, and Cheng]{liu2018bi}
Liu, Z., Wu, B., Luo, W., Yang, X., Liu, W., and Cheng, K.-T.
\newblock Bi-real net: Enhancing the performance of 1-bit {CNNs} with improved
  representational capability and advanced training algorithm.
\newblock In \emph{ECCV}, pp.\  722--737, 2018.

\bibitem[Maddison et~al.(2017)Maddison, Mnih, and Teh]{maddison2016concrete}
Maddison, C.~J., Mnih, A., and Teh, Y.~W.
\newblock The concrete distribution: A continuous relaxation of discrete random
  variables.
\newblock In \emph{ICLR}, 2017.

\bibitem[Mart{\'{\i}}nez et~al.(2020)Mart{\'{\i}}nez, Yang, Bulat, and
  Tzimiropoulos]{Martinez20}
Mart{\'{\i}}nez, B., Yang, J., Bulat, A., and Tzimiropoulos, G.
\newblock Training binary neural networks with real-to-binary convolutions.
\newblock In \emph{ICLR}, 2020.

\bibitem[Meng et~al.(2020)Meng, Bachmann, and Khan]{meng-20}
Meng, X., Bachmann, R., and Khan, M.~E.
\newblock Training binary neural networks using the {Bayesian} learning rule.
\newblock In \emph{ICML}, 2020.

\bibitem[Mnih \& Gregor(2014)Mnih and Gregor]{Mnih-2014}
Mnih, A. and Gregor, K.
\newblock Neural variational inference and learning in belief networks.
\newblock In \emph{ICML}, volume~32 of \emph{{JMLR Proceedings}}, pp.\
  1791--1799, 2014.

\bibitem[{\~{N}}anculef et~al.(2020){\~{N}}anculef, Mena, Macaluso, Lodi, and
  Sartori]{Nanculef-20}
{\~{N}}anculef, R., Mena, F.~A., Macaluso, A., Lodi, S., and Sartori, C.
\newblock Self-supervised bernoulli autoencoders for semi-supervised hashing.
\newblock \emph{CoRR}, abs/2007.08799, 2020.

\bibitem[O'Donnell(2014)]{Donnell-14}
O'Donnell, R.
\newblock \emph{Analysis of Boolean Functions}.
\newblock Cambridge University Press, USA, 2014.
\newblock ISBN 1107038324.

\bibitem[Pervez et~al.(2020)Pervez, Cohen, and Gavves]{Pervez-20}
Pervez, A., Cohen, T., and Gavves, E.
\newblock Low bias low variance gradient estimates for boolean stochastic
  networks.
\newblock In \emph{ICML}, volume 119, pp.\  7632--7640, 13--18 Jul 2020.

\bibitem[Peters \& Welling(2018)Peters and Welling]{peters2018probabilistic}
Peters, J.~W. and Welling, M.
\newblock Probabilistic binary neural networks.
\newblock \emph{arXiv preprint arXiv:1809.03368}, 2018.

\bibitem[Raiko et~al.(2015)Raiko, Berglund, Alain, and Dinh]{Raiko-14}
Raiko, T., Berglund, M., Alain, G., and Dinh, L.
\newblock Techniques for learning binary stochastic feedforward neural
  networks.
\newblock In \emph{ICLR}, 2015.

\bibitem[Rastegari et~al.(2016)Rastegari, Ordonez, Redmon, and
  Farhadi]{rastegari2016xnor}
Rastegari, M., Ordonez, V., Redmon, J., and Farhadi, A.
\newblock {XNOR-Net}: Imagenet classification using binary convolutional neural
  networks.
\newblock In \emph{ECCV}, pp.\  525--542. Springer, 2016.

\bibitem[Roth et~al.(2019)Roth, Schindler, Fr{\"o}ning, and Pernkopf]{Roth-19}
Roth, W., Schindler, G., Fr{\"o}ning, H., and Pernkopf, F.
\newblock Training discrete-valued neural networks with sign activations using
  weight distributions.
\newblock In \emph{European Conference on Machine Learning (ECML)}, 2019.

\bibitem[Shayer et~al.(2018)Shayer, Levi, and Fetaya]{shayer2018learning}
Shayer, O., Levi, D., and Fetaya, E.
\newblock Learning discrete weights using the local reparameterization trick.
\newblock In \emph{ICLR}, 2018.

\bibitem[Shekhovtsov \& Yanush(2021)Shekhovtsov and Yanush]{ST}
Shekhovtsov, A. and Yanush, V.
\newblock Reintroducing straight-through estimators as principled methods for
  stochastic binary networks.
\newblock In \emph{GCPR}, 2021.

\bibitem[Shekhovtsov et~al.(2020)Shekhovtsov, Yanush, and Flach]{PSA}
Shekhovtsov, A., Yanush, V., and Flach, B.
\newblock Path sample-analytic gradient estimators for stochastic binary
  networks.
\newblock In \emph{NeurIPS}, 2020.

\bibitem[Shen et~al.(2018)Shen, Su, Chapfuwa, Wang, Wang, Henao, and
  Carin]{Shen-18}
Shen, D., Su, Q., Chapfuwa, P., Wang, W., Wang, G., Henao, R., and Carin, L.
\newblock {NASH}: Toward end-to-end neural architecture for generative semantic
  hashing.
\newblock In \emph{Annual Meeting of the Association for Computational
  Linguistics}, 2018.

\bibitem[Tang et~al.(2017)Tang, Hua, and Wang]{Tang2017HowTT}
Tang, W., Hua, G., and Wang, L.
\newblock How to train a compact binary neural network with high accuracy?
\newblock In \emph{AAAI}, 2017.

\bibitem[Tucker et~al.(2017)Tucker, Mnih, Maddison, Lawson, and
  Sohl-Dickstein]{Tucker-17-REBAR}
Tucker, G., Mnih, A., Maddison, C.~J., Lawson, J., and Sohl-Dickstein, J.
\newblock {REBAR}: Low-variance, unbiased gradient estimates for discrete
  latent variable models.
\newblock In \emph{NeurIPS}, 2017.

\bibitem[Vahdat et~al.(2020)Vahdat, Andriyash, and Macready]{Vahdat2020PMLR}
Vahdat, A., Andriyash, E., and Macready, W.
\newblock Undirected graphical models as approximate posteriors.
\newblock In \emph{ICML}, volume 119, pp.\  9680--9689, 13--18 Jul 2020.

\bibitem[Xiang et~al.(2017)Xiang, Qian, and Yu]{Xiang-17}
Xiang, X., Qian, Y., and Yu, K.
\newblock Binary deep neural networks for speech recognition.
\newblock In \emph{INTERSPEECH}, 2017.

\bibitem[Yin \& Zhou(2019)Yin and Zhou]{yin18-arm}
Yin, M. and Zhou, M.
\newblock {ARM}: Augment-{REINFORCE}-merge gradient for stochastic binary
  networks.
\newblock In \emph{ICLR}, 2019.

\bibitem[Zhou et~al.(2016)Zhou, Wu, Ni, Zhou, Wen, and Zou]{zhou2016dorefa}
Zhou, S., Wu, Y., Ni, Z., Zhou, X., Wen, H., and Zou, Y.
\newblock Dorefa-net: Training low bitwidth convolutional neural networks with
  low bitwidth gradients.
\newblock \emph{arXiv preprint arXiv:1606.06160}, 2016.

\end{thebibliography}
%\printbibliography
%\putbib
%\end{bibunit}
}

\newpage
\newpage
\onecolumn
%\newgeometry{left=2cm, righ=2cm,top=2cm,bottom=2cm}
%\newgeometry{includefoot,margin=2cm,bottom=1in}
\appendix
\numberwithin{figure}{section}
\addcontentsline{toc}{chapter}{Appendix}
%\addtocontents{toc}{\protect\setcounter{tocdepth}{2}}
%\pagestyle{plain}
%
% Reset counters
\setcounter{figure}{0}
\setcounter{table}{0}
\counterwithin{figure}{section}
\counterwithin{table}{section}
\counterwithin{theorem}{section}
\counterwithin{proposition}{section}
\counterwithin{lemma}{section}
\counterwithin{corollary}{section}
%
%\title{\mytitle (Appendix)}
%\author{Paper ID \SubNumber \\[10pt]}
%\author{ACML Submission}
%\title{Appendix}
%\author{}
%\maketitle
%
%%\appendices
\addtocontents{toc}{\protect\setcounter{tocdepth}{2}}
%\pagestyle{plain}
%
%% Reset counters
%\setcounter{figure}{0}
%\setcounter{table}{0}
%\counterwithin{figure}{section}
%\counterwithin{table}{section}
%
%\twocolumn[{%
 %\centering
 %\LARGE \mytitle \\ %Supplementary Material \\[1.5em] 
 %(CVPR Submission \#\cvprPaperID\ Supplementary Material) \\[1.5em]
 %\normalsize
%}]
%
% TOC
%remove dots and page numbers
\let\Contentsline\contentsline
\renewcommand\contentsline[3]{\Contentsline{#1}{#2}{}}
\makeatletter
\renewcommand{\@dotsep}{10000}
\makeatother

\let\authcount\relax
\def\authcount#1{}
{\centering
 \LARGE %\mytitle %Supplementary Material \\[1.5em] 
  Appendix \\[1.5em]
\normalsize
}
%
%\tableofcontents

%\begin{refsection}
%
%\defaultbibliographystyle{abbrvnat}
%\setcitestyle{authoryear,open={((},close={))},round}
%\setcitestyle{notesep={; },round,aysep={},yysep={;}}
%\setcitestyle{authoryear}
%\begin{bibunit}
%
%\input{tex/related.tex}
%\input{tex/more-ST.tex}
%\input{tex/more-md.tex}
%\input{tex/more-VB.tex}
%\input{tex/more-exp.tex}
%\clearpage
\section{Gumbel Softmax and ST Gumbel-Softmax}\label{sec:GSA}

%\begin{restatable}{proposition}{PropGS1}\label{PropGS1}
%The estimator is asymptotically unbiased as $\tau \rightarrow 0$ and the bias decreases at the rate $O(\tau)$.
%\end{restatable}
\PropGS*
\begin{proof}
%By standard arguments for the expected value of the mean, this is equivalent to showing that 
%\begin{align}
%\lim_{t\rightarrow 0+} \hat f'_t(x) = f'(x),
%\end{align}
%where
Let us denote
\begin{align}
\bar g_\tau = \E_z [ \ggst] = \int_{-\infty}^{\infty} \frac{d}{d \eta} \f (\sigma_\tau(\eta - z)) p_z(z) \d z.
\end{align}
Note that $\lim_{\tau \rightarrow 0+} \bar g_\tau(x)$ cannot be simply be evaluated by moving the limit under the integral --- no qualification theorem allows this. We apply the following reformulation. The derivative $\frac{d}{d \eta} \f(\sigma_\tau(\eta - z))$ expands as
\begin{align}
\f'(\sigma_\tau(\eta-z)) \sigma_\tau(\eta-z) (1-\sigma_\tau(\eta-z)) \frac{1}{\tau}.
\end{align}
We make a change of variables $v = \sigma_\tau(\eta-z)$ in the integral. This gives $z = \eta - \tau \logit(v)$ and $\d z = -\tau \frac{1}{v(1-v)} \d v$. Substituting and cancelling part of the terms, we obtain
\begin{align}\label{concrete-deriv}
\bar g_\tau = \int_{0}^{1} \f'(v) p_z(\eta - \tau \logit(v)) \d v.
\end{align}
With this expression we can now interchange the limit and the integral using the dominated convergence theorem.
In order to apply it we need to show that there exist an integrable function $h(v)$ such that
\begin{align}
|\f'(v)p_z(\eta - \tau \logit(v))| < h(v)
\end{align}
for all $\tau>0$.
Observe that
\begin{align}
\sup_{v \in [0,1]}|p_z(\eta - \tau \logit(v))|  = \sup_{u \in \Real}|p_z(\eta - \tau u)| = \sup_{y \in \Real}|p_z(y)| = p_z(0) = \frac{1}{4},
\end{align}
where we used that the maximum of standard logistic density is attained at zero.
We can therefore let $h(v) = \f'(v)/4$. Since $\f'(v)$ is the derivative of $\f$, it is integrable on $[0,1]$.
Therefore the conditions of the dominated convergence theorem are satisfied and we have
\begin{subequations}
\begin{align}\label{concrete-deriv-limit}
\lim_{\tau\rightarrow 0+}\bar g_\tau & = \int_{0}^{1} \f'(v) \lim_{\tau\rightarrow 0+}  p_z(\eta - \tau \logit(v)) \d v\\
& = \int_{0}^{1} \f'(v) p_z(\eta) \d v =  (\f(1) - \f(0)) p_z(\eta),
\end{align}
\end{subequations}
which is the correct value of the gradient in $\eta$.

Next, we obtain the series representation of the estimator bias in the asymptote $\tau\rightarrow 0$.
We approximate $p_z(\eta - \tau \logit(v))$ with its Taylor series around $\tau=0$:
\begin{align}
p_z(\eta - \tau \logit(v)) = p_z(\eta)\Big(1  + c_1 \logit(v) \tau + c_2 \logit^2(v) \tau^2 + O(\tau^3) \Big),
\end{align}
where 
\begin{align}
c_1 = \frac{e^\eta-1}{e^\eta+1};\ \ \ \ \
c_2 = \frac{-4 e^\eta+e^{2 \eta}+1}{2 (e^\eta+1)^2}.
\end{align}
This is obtained using Mathematica. We use this expansion in the integral representation~\cref{concrete-deriv}. Observing that $\int_{0}^{1} \f'(v) \d v = (\f(1)-\f(0))$ is the true gradient, the zero order term becomes the true gradient.
It follows that the bias of $\ggst$ is asymptotically
\begin{align}\label{GS-bias-expansion}
p_z(\eta)\Big( c_1 \Big(\int_{0}^{1} \f'(v) \logit(v) \d v \Big) \tau + c_2 \Big(\int_{0}^{1} \f'(v) \logit ^2(v) \d v\Big)  \tau^2\Big) + O(\tau^3).
\end{align}
In the case when $\f$ is linear, the first order term vanishes because $\f'$ is constant and $\logit(v)$ is odd about $\tfrac{1}{2}$. However $\tau^2$ and higher order even terms do not vanish, therefore the estimator is still biased even for linear objectives.
\end{proof}

\PropGSV*
\begin{proof}

We will show that the second moment of the estimator $\ggst$ has the following asymptotic expansion for $\tau\rightarrow 0$:
\begin{align}
\notag p_z(\eta	)\Big[ \Big(\int_{0}^{1} \f'(v)^2 v(1-v) \Big)\frac{1}{\tau} + &	c_1 \Big(\int_{0}^{1} \f'(v)^2 v (1-v) \logit(v) \d v \Big) \\
 + & c_2 \Big(\int_{0}^{1} \f'(v)^2 v (1-v) \logit^2(v) \d v \Big) \tau \Big] + O(\tau^2).\label{GS-second-moment}
\end{align}

The second moment expresses as
\begin{subequations}
\begin{align}
\E[\ggst^2] & = \int_{-\infty}^{\infty} (\frac{\partial}{\partial \eta} \f(\sigma_\tau(\eta - z)))^2 p_z(z) \d z\\
 & = \int_{-\infty}^{\infty} \Big(\f'(\sigma_\tau(\eta-z))\sigma_\tau(\eta-z)(1-\sigma_\tau(\eta-z))\frac{1}{\tau}\Big)^2 p_z(z)dz
\end{align}
\end{subequations}
We perform the same substitution of variables: $v = \sigma_t(\eta-z)$, $dv = -v(1-v)\frac{1}{\tau}\d z$ to obtain
%\begin{subequations}
\begin{align}\label{GS-second}
\E[\ggst^2] %& = \int_0^1 \f'(v)^2v(1-v)\frac{1}{\tau}p_z(\eta - \tau \logit(v))\d v \\
 =\frac{1}{\tau} \int_0^1 \f'(v)^2 v(1-v) p_z(\eta - \tau \logit(v))\d v.
\end{align}
%\end{subequations}
We perform the same Taylor expansion for $p_z(\eta - \tau \logit(v))$ around $\tau=0$ as in \cref{PropGS1} and combine the terms to obtain the expansion~\eqref{GS-second-moment}.
The variance is dominated by the $O(\tfrac{1}{\tau})$ term of the second moment.
\end{proof}

\PropGSB*
\begin{proof}
Considering $z$ fixed and denoting $x= \eta -z$, we need to check the asymptotic behavior of 
\begin{align}
\frac{\d }{\d x}\sigma_\tau(x) = \frac{1}{\tau}p_z(x/\tau) = \frac{e^{-\frac{x}{\tau}}}{\tau (1 + e^{-x/\tau})^2}
\end{align}
as $\tau \rightarrow 0$. Since $p_z$ is symmetric, we may assume $x>0$ without loss of generality. The denominator is then asymptotically just $\tau$. Therefore the ratio is asymptotically $O(\frac{1}{\tau} c^{1/\tau})$. % with $c=e^{|x|} \geq 1$.
\end{proof}

\PropGSC*

\begin{proof}
We want to analyze the probability
\begin{align}
P = \Pr(\frac{\d}{\d \eta}\sigma_\tau(\eta-z) \geq \varepsilon )
\end{align}
when $z$ is distributed logistically. Let $s = \sigma_\tau(\eta-z)$. Then $\frac{\d}{\d \eta} \sigma_\tau(\eta-z) = s(1-s)$. The equality $s(1-s) = \varepsilon$ holds for $s^* = \tfrac{1}{2}(1 - \sqrt{1-4\eps})$. This implies
\begin{align}
z_{1,2} = \eta \pm \tau \logit(s^*).
\end{align}
The inequality $s(1-s) \geq \varepsilon$ holds in the interval $[z_1, z_2]$. Thus the probability in question is given by
\begin{align}\label{F-diff-tau}
P = \sigma(z_2) - \sigma(z_1).
\end{align}
As $\tau \logit(s^*) \rightarrow 0$ for $\tau\rightarrow 0$, we have asymptotically that
\begin{align}
P = p_z(\eta) \tau \logit(s^*).
\end{align}
Finally, note that $\logit(\tfrac{1}{2}(1 - \sqrt{1-4\eps}))$ is asymptotically $O(-\log \varepsilon)$ for $\varepsilon\rightarrow 0$.
%Because of the scaling by $\tau$ the probability to obtain a gradient of magnitude at least $\varepsilon$ scales at $O(\tau \log \frac{1}{\varepsilon})$ as $\tau \rightarrow 0$ and $\varepsilon \rightarrow 0$. 
\end{proof}

%\begin{restatable}{proposition}{PropGS4}\label{PropGS4}
%The variance of GS estimator grows as $O(\frac{1}{\tau})$.
%\end{restatable}

\PropSTGS*
\begin{proof}
Notice that the derivative $f'(x)$ in ST-GS only takes two values: $f'(1)$ for $\eta-z \geq 0$ and $f'(0)$ otherwise. Introducing $v = \tau \sigma_\tau(\eta -z)$, the condition $\eta \geq z$ can be equivalently written as $\eta \geq \eta - \tau \logit(v)$, which reduces to $v \leq 0.5$. In the expected GS gradient value we can therefore substitute
\begin{align}\label{STGS-subst}
\begin{cases}
f'(v) = f'(1) & \text{\ if \ \ } v \leq 0.5,\\
f'(v) = f'(0) & \text{\ otherwise},
\end{cases}
\end{align}
and obtain for the expected value of the estimate
\begin{align}
\bar g_\tau = f'(1) \int_{0}^{0.5} p_z(\eta - \tau \logit(v))\d v + f'(0) \int_{0.5}^{1} p_z(\eta - \tau \logit(v))\d v.
\end{align}
Substituting the Taylor expansion of $p_z(\eta - \tau \logit(v))$ in $\tau$, we obtain
\begin{align}
\bar g_\tau = p_z(\eta) \Big((f'(1) + f'(0))\frac{1}{2} + (f'(0) - f'(0)) c_1 \log(2) \tau \Big) + O(\tau^2).
\end{align}
We see that in the asymptote $\tau\rightarrow 0$, the expected value of the estimator approaches  $p_z(\eta) (f'(1) + f'(0))\frac{1}{2}$, which matches the DARN estimator (confer to~\eqref{E-DARN}, which is the expected DARN gradient in $p$ for $\{-1,1\}$ variables, while $g_\tau$ is in $\eta$ and $\{0,1\}$ variables). Therefore, bias, approaches that of DARN, in particular the bias vanishes for quadratic $f$. %The asymptotic gradient $p_z(\eta) 0.5 (f'(1) + f'(0))$ is correct for quadratic functions. 
Indeed, let $f(x) = a x^2 + b x +c$, then the correct gradient is $p_z(\eta)(a + b)$.

The variance of ST GS estimator is dominated by the second moment, which has asymptotic expansion~\eqref{GS-second} with the substitution~\eqref{STGS-subst}. The variance is dominated by the $O(\tfrac{1}{\tau})$ term of the second moment. More specifically we have the asymptote of the second moment
\begin{align}
\frac{1}{\tau}\Big(f'(1)^2 + f'(0)^2\Big)\frac{1}{12} +  \Big(f'(1)^2 - f'(0)^2\Big)\frac{1-4 \log(2)}{24} + O(\tau).
\end{align}
%
%
%
%
%We obtain the expansion
%\begin{align}\label{GS-bias-expansion}
%c_1 \log(2) (f'(0) - f'(1)) \tau + c_2(x) \Big(\int_{0}^{1} \f'(v) \logit ^2(v) \d v\Big)  \tau^2 + O(\tau^3).
%\end{align}
%
%
%Since $f'(x)$ does not vary with $z$ unless $x=\leftbb \eta-z \geq \rightbb$ changes, to compute the expected gradient we can marginalize over $z$ separately in the two regions: $z < \eta$ and $z > \eta$. Let's consider the first case. We have:
%\begin{align}
%\int_{-\infty}^{\eta} \frac{1}{\tau}\sigma_\tau(\eta - z) (1 - \sigma_\tau(\eta - z))p_z(z) \d z.
%\end{align}
%We make a substitution $z = \eta - \tau \logit(v)$. Then $v = \sigma_\tau(\eta-z)$ and $\d z = -\tau \frac{1}{v(1-v)} \d v$. The integral expresses as
%\begin{align}
%-\int\limits_{1}^{0.5} p_z(\eta - \tau \logit(v)) \d v.
%\end{align}
%The limit of this integral with $\tau\rightarrow 0$ is $\frac{1}{2}p_z(\eta)$. And the same holds for the case $x = 0$, \ie, $z>\eta$. We obtained, is that ST Gumbel-Softmax in the simple case of linear objective and in the favorable limit $\tau\rightarrow 0$ underestimates the true gradient by the factor $1/2$.
\end{proof}
\section{BayesBiNN}\label{sec:BayesBiNNA}

\PropGStanh*
\begin{proof}
This result is already proven in \citet[Lemma 1]{meng-20}. We nevertheless repeat the derivation in order to make sure there is no mismatch due to a different notation and in order to expand proof details omitted in~\cite{meng-20}.

Recall that $\tanh(\lambda) = 2\sigmoid(2 \lambda) - 1$.
Let us define $x = (w+1)/2$, $p = (\mu+1)/2$, $\tilde f\colon \{0,1\}\mapsto\Real\colon x\to f(2 x-1)$ and $\eta = 2\lambda$. We have $x \sim \Bernoulli(p)$, $p = \sigmoid(\eta)$ and $\E_w[f(w)] = \E_x[\tilde f(x)] =: E$.
According to~\eqref{GS-grad}, the \GS estimate of gradient of $E$ in $\eta$ is given by
\begin{subequations}
\begin{align}
z & \sim \text{Logisitc},\\
\tilde x & = \sigma_\tau(\eta - z),\\
\frac{\hat \d E} {d \eta} & = \tilde f'(\tilde x) \frac{\partial \tilde x}{\partial \eta}.
\end{align}
\end{subequations}
Defining $\tilde w = 2\tilde x-1$ and back-substituting, we obtain
\begin{subequations}
\begin{align}
\tilde w &  = \tanh_\tau(\frac{\eta - z}{2}) = \tanh_\tau(\lambda - z/2),\\
\frac{\hat \d E} {d \lambda} = 2 \frac{\hat \d E} {d \eta} & = 2 \Big(2 f'(\tilde w)\Big) \Big(\frac{1}{4}\frac{\partial \tilde w}{\partial \lambda}\Big) = f'(\tilde w)\frac{\partial \tilde w}{\partial \lambda}.
\end{align}
\end{subequations}
Respectively, \GS estimator of gradient $\frac{\d \E_w[f]}{\d \mu}$ is given by
\begin{align}
\hat g = f'(\tilde w) \frac{\partial \tilde w}{\partial \lambda}\Big(\frac{\partial \mu}{\partial \lambda}\Big)^{-1}.
\end{align}
Using the identity $\frac{\d }{\d x}\tanh(x) = 1-\tanh^2(x)$, we obtain
\begin{align}
\hat g = f'(\tilde w) \frac{1 - \tilde w^2}{\tau(1 - \mu^2)}.
\end{align}
Finally note that $z/2 \sim \frac{1}{2}\text{Logistic}$.
\end{proof}

\PropBayesBiNNA*
\begin{proof}
%For simplicity we will assume that $\eta$ and $g$ as scalars. For the general vector case the arguments would apply coordinate-wise.

First, we analyze the nominator of J~\eqref{BayesBiNN-J}. From the asymptotic expansion of
\begin{align}
1 - \tanh[\log(x)]^2 = \frac{4}{x^2}+O(\frac{1}{x^3}), \text{for\ } x\rightarrow \infty,
\end{align}
substituting $\log(x) = \frac{|\lambda-\delta|}{\tau}$, we obtain
\begin{align}
1 - \tilde w^2 \sim 4 \exp\big(-2\frac{|\lambda-\delta|}{\tau}\big).
\end{align}
% Observe that $1 - w_b^2 = \Big(\tanh(\frac{\eta-z}{\tau})\Big)^2$ is asymptotically $4 \Big(e^{-\frac{\eta-z}{\tau}}\Big)^2$.
%scales as $c^{\frac{2}{\tau}}$ with $c = e^{-|\eta-z|} \leq 1$. 
For example, for $\lambda = 5$ and $\delta=0$ we have that $(1 - \tilde w^2) \approx 4 e^{-10^{11}}$. Therefore in the nominator, the part $(1 - \tilde w^2)$ is negligibly small compared to $\eps$ and even to the floating point precision. This applies so long as $|\lambda-\delta| \gg  \tau = 10^{-10}$, which we expect to hold with high probability for two reasons: 1) $\lambda$ will be shown to grow significantly during the first iterates and 2) the probability of the noise $\delta$ matching $\lambda$ to this accuracy even for $\lambda=0$ is of the order $O(\tau)$.

The denominator of J~\eqref{BayesBiNN-J} satisfies the bounds
\begin{align}
\tau\epsilon \leq \tau (1-\tanh(\lambda)^2 + \epsilon) \leq \tau (1+\epsilon),
\end{align}
from which we can conclude that $J \geq \frac{\eps}{\tau} = 1$. However for a moderately large $\lambda$ the denominator drops quickly, \eg for $\lambda=5$, we have $\tau(1-\tanh(\lambda)^2 +\eps) < \tau (2\cdot 10^{-4} +\eps) \approx 2\cdot 10^{-4} \tau$. And the decrease rate for $|\lambda |\rightarrow \infty$ is $\tau (4 e^{-2|\lambda|} + \epsilon)$.

Since $\lambda$ is initialized uniformly in $[-10,10]$ and receives updates of order at least $\alpha N f' \approx 5 f'$ (for the initial $\alpha=10^{-4}$ used), during the first iterates $|\lambda|$ can be expected to grow significantly until we reach the asymptote $e^{-2|\lambda|} \ll \eps$, which is when $|\lambda| > 5 \log 10 \approx 11$. After reaching this asymptote, we will have $J \approx \frac{ \epsilon}{\tau\epsilon} = \frac{1}{\tau}$ and we may expect the growth of $\lambda$ to stabilize around $|\lambda| \approx \alpha \frac{N}{\tau} |f'| \approx 10^{10}$.

The first consequence of this is that the scaling factor that was supposed to implement Gumbel-Softmax gradient, just becomes an inadvertent constant $\frac{1}{\tau}$. %It is tempting to conclude that this up-scaling of the gradient corresponds to solving the variational Bayesian problem with the data evidence part up-scaled respectively, \ie, completely dominating the the KL prior part, but this is not exactly so because also the true stochastic scaling of the gradient is modified. 

The second consequence is that the natural parameters $\lambda$ have huge magnitudes during the training, and we have that $|\delta| \ll |\lambda|$ with high probability, therefore the noise plays practically no role even in the forward pass of BayesBiNN. In this mode the BayesBiNN algorithm becomes equivalent to
\begin{subequations}\label{BayesBiNN-step-R1A}
\begin{align}
w &:= \sign(\lambda);\\
\lambda &:= (1-\alpha) \lambda - \alpha \frac{N}{\tau} f'(w).\label{BayesBiNN-step}
\end{align}
\end{subequations}
It is seen that the forward pass and the gradient implement the deterministic straight-through with identity derivative and that the update has a form of SGD with a latent weight decay and with the gradient of data evidence up-scaled by $\frac{1}{\tau}$. These huge step-sizes of $O(\frac{1}{\tau})$ do not destroy the learning because $\sign$ is invariant to a global rescaling of $\lambda$. 

Denoting $\bar \lambda = \frac{\tau}{N} \lambda$, we can equivalently rewrite~\eqref{BayesBiNN-step-R1A} as 
\begin{subequations}
\begin{align}\label{BayesBiNN-det}
w &:= \sign(\bar \lambda);\\
%\vg &:= \nabla_{\vw_b} \L;\\
\bar \lambda &:= (1-\alpha) \bar \lambda - \alpha f'(w).
\end{align}
\end{subequations}
This algorithm and the resulting binary weights $w$ do not depend on $\tau$, $N$.
\end{proof}
\section{FouST}\label{sec:FouSTA}
\PropFouSTNosie*
\begin{proof}
%Following the notation of ~\cite{Pervez-20}, let $\rho \in[0,1]$ and let $x'\sim N_\rho(x)$ denote that $x'$ is set equal to $x$ with probability $\rho$ and set to an independent sample from $\Bin(p)$ with probability $1-\rho$.
%Indeed, the Fourier noise operator $T\rho[f](x) = \E_{x'\sim N_\rho(x)[f(x')]}$~\cite[Sec. 2]{Pervez-20} reduces both bias and variance, ultimately to zero when $\rho=0$, \ie samples $x'$ are independent of $x$. However, the method proposed in~\cite[Sec. 4.2]{Pervez-20} approximates this expectation using $S$ samples $x^{(s)}\sim N_\rho(x)$ as follows: %. The estimator that is proposed is defined as
%\begin{align}
%\hat g_{\rho} = \frac{1}{S} \sum_{s} \hat g (x^{(s)}),
%\end{align}
%where $\hat g$ is the base \ST or \DARN estimator. 
We have
\begin{subequations}
\begin{align}
\E[\hat g_{\rho}] & = \E_x \Big[ \frac{1}{S} \sum_{s} \E_{x'\sim N_\rho(x)}[\hat g (x')] \Big] \\ 
& = \E_x \Big[ \frac{1}{S} \sum_{s} (\rho \hat g (x) + (1-\rho)\E_{x'} [\hat g (x')]) \Big] = \E[\hat g] =: \mu,
\end{align}
\end{subequations}
where $\E_{x'}$ denotes samples from the base distribution $\Bin(p)$.
Therefore this estimator, unlike the motivating operator $T_\rho[f](x)$, does not reduce the bias, no matter how many samples we use. 
%We now compare 
%\begin{align}
%& \V[\hat g_{\rho}] - \V[\hat g_{0}] = \E[\hat g_{\rho}^2] - \E[\hat g_{0}^2]\\
%& = \E\big[\big(\frac{1}{S} \sum_{s} \hat g (x^{(s)})\big)^2\big] - \frac{1}{S} \E[\hat g (x)^2]\\
%& = \frac{1}{S^2} \E\big[\sum_{s} \hat g (x^{(s)})^2 \big] + \frac{1}{S^2} \E\big[\sum_{s\neq t} \hat g (x^{(s)}) \hat g (x^{(t)})\big] - \frac{1}{S} \E[\hat g (x)^2]\\
%& = \frac{1}{S^2} \E_{x,x',x''}\big[\sum_{s\neq t} (\rho \hat g (x) + (1-\rho) \hat g (x')) (\rho \hat g (x) + (1-\rho) \hat g (x'')) \big]\\
%& = \frac{1}{S^2} \E_{x,x',x''}\big[\sum_{s\neq t} (\rho^2 \hat g (x)^2 + 2 \rho(1-\rho) \hat g (x')\hat g (x) + (1-\rho)^2 \hat g (x')\hat g (x'')) \big]\\
%& = \frac{S-1}{S}\Big( \rho^2 \E_x\big[( \hat g (x)^2]  + ( 2 \rho(1-\rho) + (1-\rho)^2) (\E_z [\hat g (x)])^2\Big)\\
%& = \frac{S-1}{S}\Big( \rho^2 \E_x\big[( \hat g (x)^2]  + ( 1 - \rho^2) (\E_z [\hat g (x)])^2 \Big)\\
%%\frac{1}{S^2} \E\big[\sum_{s,t} \hat g (x^{(s)}) \hat g (x^{(t)})\big] - \frac{1}{S} \E[\hat g (x)^2]=\\
%%\frac{1}{S^2} \big[\sum_{s,t} \E_{x, x^{(s)}}(\rho \hat g (x) + (1-\rho)\E_{x'} [\hat g (x')]) \hat g (x^{(t)})\big] - \frac{1}{S} \E[\hat g (x)^2]=\\
%\end{align}
We can express the variance as
\begin{align}
& \V[\hat g_{\rho}] = \E[\hat g_{\rho}^2] - \mu^2\\
& = \E\big[\big(\frac{1}{S} \sum_{s} \hat g (x^{(s)})\big)^2\big] - \E[\hat g (x)^2]\\
& = \frac{1}{S} M + \frac{1}{S^2} \E\big[\sum_{s\neq t} \hat g (x^{(s)}) \hat g (x^{(t)})\big] - \mu^2,
\end{align}
where $M = \E\big[\hat g (x)^2 \big]$ and further expand
\begin{align}
& \frac{1}{S^2} \E\big[\sum_{s\neq t} \hat g (x^{(s)}) \hat g (x^{(t)})\big] = \\
& \frac{1}{S^2} \E_{x,x',x''}\big[\sum_{s\neq t} (\rho \hat g (x) + (1-\rho) \hat g (x')) (\rho \hat g (x) + (1-\rho) \hat g (x'')) \big]\\
& = \frac{1}{S^2} \E_{x,x',x''}\big[\sum_{s\neq t} (\rho^2 \hat g (x)^2 + 2 \rho(1-\rho) \hat g (x')\hat g (x) + (1-\rho)^2 \hat g (x')\hat g (x'')) \big]\\
& = \frac{S-1}{S}\Big( \rho^2 \E_x\big[( \hat g (x)^2]  + ( 2 \rho(1-\rho) + (1-\rho)^2) (\E_z [\hat g (x)])^2\Big)\\
& = \frac{S-1}{S}\Big( \rho^2 \E_x\big[( \hat g (x)^2]  + ( -2 \rho^2 + 1 + \rho^2) (\E_z [\hat g (x)])^2 \Big)\\
& = \frac{S-1}{S}\Big( \rho^2 M  + ( 1 - \rho^2) \mu^2 \Big).
%\frac{1}{S^2} \E\big[\sum_{s,t} \hat g (x^{(s)}) \hat g (x^{(t)})\big] - \frac{1}{S} \E[\hat g (x)^2]=\\
%\frac{1}{S^2} \big[\sum_{s,t} \E_{x, x^{(s)}}(\rho \hat g (x) + (1-\rho)\E_{x'} [\hat g (x')]) \hat g (x^{(t)})\big] - \frac{1}{S} \E[\hat g (x)^2]=\\
\end{align}
In total we obtain:
\begin{align}
& \V[\hat g_{\rho}] = \frac{1}{S}\Big((1 + (S-1)\rho^2) M + ((S-1)(1 - \rho^2) - S)\mu^2 \Big).
\end{align}
Respectively,
\begin{align}
& \V[\hat g_{\rho}] -\V[\hat g_{0}] = \frac{S-1}{S}\rho^2 \sigma^2.
%&= \rho^2 M + (\frac{S-1}{S} (-2 \rho^2 + 1 + \rho^2) - \frac{1}{S})\mu^2  \\
\end{align}
This shows that variance using correlated samples is always higher.
\end{proof}

\end{document}